\pdfoutput=1
\documentclass[11pt]{article}

\usepackage[preprint]{ACL2023}

\usepackage{times}
\usepackage{latexsym}

\usepackage[T1]{fontenc}
\usepackage[utf8]{inputenc}
\usepackage{booktabs}       % professional-quality tables

\usepackage{microtype}

\usepackage{graphicx}

\usepackage{inconsolata}

\title{Reconsidering Sentence-Level Sign Language Translation}

\author{%
  Garrett Tanzer$^1$\thanks{~~Correspondence to \texttt{gtanzer@google.com}.}~, Maximus Shengelia$^2$\thanks{~~Work done while at Google.}~, Ken Harrenstien$^1$, David Uthus$^1$ \\
  $^1$Google, $^2$Rochester Institute of Technology}

\begin{document}
\maketitle
\begin{abstract}
Historically, sign language machine translation has been posed as a sentence-level task: datasets consisting of continuous narratives are chopped up and presented to the model as isolated clips. In this work, we explore the limitations of this task framing. First, we survey a number of linguistic phenomena in sign languages that depend on discourse-level context. Then as a case study, we perform the first human baseline for sign language translation that actually substitutes a human into the machine learning task framing, rather than provide the human with the entire document as context. This human baseline---for ASL to English translation on the How2Sign dataset---shows that for 33\% of sentences in our sample, our fluent Deaf signer annotators were only able to understand key parts of the clip in light of additional discourse-level context. These results underscore the importance of understanding and sanity checking examples when adapting machine learning to new domains.
\end{abstract}

\section{Introduction}

One of the key challenges in sign language processing is that methods from mainstream natural language processing (NLP) are tailored primarily to text and secondarily to speech. Much of the work in this space therefore focuses on generalizing these methods to video, in order to capture this oft-neglected dimension of linguistic diversity \citep{bragg, kayo}.

One such carryover is that sign language machine translation (MT) is framed as a sentence-level task. Although continuous sign language datasets are usually derived from long-form signed content (e.g., interpreted news broadcasts), they are preprocessed into short clips associated with each sentence in the spoken language transcript (which may not themselves correspond to discrete sentences in the continuously translated sign language version), and models are trained and evaluated on these clips in isolation. In this work, we examine the limitations of this task framing, which---like many other sign language modeling decisions~\citep{desai2024systemic}---was adopted somewhat uncritically, and ask: what is the right unit of translation for sign language?

Machine translation between spoken languages is typically posed as a sentence-level task, and although it largely works, there are known intersentential dependencies like anaphora that are impossible to resolve in isolation~\citep{bawden-etal-2018-evaluating,spokenContext}. These dependencies are especially troublesome for language pairs that have mismatches in grammatical features like pronoun dropping, tense marking, or gradations of register.

The situation is perhaps even more pronounced for translation between spoken languages and sign languages. Sign languages are not just spoken languages produced with the hands: the grammar of sign languages is shaped by the nature of the visual-spatial modality~\citep{texas2002modality}. While utterances produced by non-native signers tend to resemble the syntax of the region's spoken language, native signing often expresses concepts in a fundamentally different way that is richly grounded in spatial world understanding and, more importantly here, the discourse context. When deprived of that context, the viewer may catastrophically fail to understand the meaning of an utterance and therefore be unable to translate it. We describe some linguistic phenomena relevant to cross-modal translation in Section~\ref{sec:linguistics}.

To the best of our knowledge, no sign language MT benchmarks provide baselines for human performance that actually ask humans to perform the same task that they expect of the model. Reference translations are given in the dataset by construction, either as the source text or by discourse-level translation. Human judgments are used at the discourse level to quality-check preprocessing or to evaluate model-generated outputs, but not to sanity check the task framing itself.

We therefore provide in Section~\ref{sec:evaluation} the first such human baseline, for American Sign Language (ASL) to English translation on the \textit{How2Sign} dataset~\citep{how2sign}, as a case study. How2Sign consists of informal instructional (``how to'') narratives, which is a particularly illustrative domain. Before even scoring results against ground truth references, we find that for 33.3\% of instances in our sample, our fluent Deaf signer annotators felt that they could not fully perform the translation given only the sentence-level clip---but could, given additional discourse-level context. Most of these errors were due to features of sign languages that lack direct analogues in spoken languages. When we do compute metrics, we get a surprisingly low score of 19.8 BLEU (56.6 BLEURT) for the sentence-level task, which increases with additional context but only to 21.5 (59.5). We disaggregate these results for each of five distinct interpreters in the How2Sign test set, and find that sentence-level results vary from 5.2 BLEU (45.7 BLEURT) to 39.5 (70.0) across individuals. Scores are higher for interpreters who hew closer to English; context is more important for those who don't.

We hope that these results and analysis will encourage the sign language MT field to reconsider whether computational benefits of the sentence-level task framing outweigh its quality and alignment limitations, and to continue to pare back unfounded modeling assumptions by understanding datasets more deeply and crafting benchmarks more deliberately.

\section{Background \& Related Work}

\subsection{Sign Languages}

See~\citet{bragg},~\citet{kayo},~\citet{decosterSurvey}, and~\citet{desai2024systemic} for excellent surveys of the social and technical aspects of sign language processing.

In brief, in contrast to spoken languages, which are articulated with the vocal tract, sign languages are articulated with the upper body (including the face). These two modalities impose different constraints on the grammar of languages within them. Sign languages are minority languages primarily used by the Deaf/Hard of Hearing communities of various regions; they are natural languages that are genealogically unrelated to but often considerably influenced by the dominant spoken language of the region. Within a single sign language, there is a great deal of variation due to geographic and social factors.

For example, in the US and Canada there is a diglossic spectrum from American Sign Language (ASL), a fully-fledged independent language; to Manually Coded English (MCE), a system used to transcribe spoken English into the sign lexicon of ASL; with Conceptually Accurate Signed English (CASE) vaguely in between~\citep{supalla2002role,rendel2018signing}. Across all of these, there is regional variation in vocabulary, analogous to ``soda'' vs. ``pop'' in American English but perhaps more pronounced~\citep{shroyer1984signs}. And there is social variation, like Black ASL, analogous to Black English~\citep{blackasl}. Less than 6\% of deaf children in the US and less than 2\% of deaf children worldwide are exposed to a sign language in early childhood~\citep{signExposurePct}, so there are also different levels of proficiency even among Deaf signers. MT should handle all these dimensions of variation.

\subsection{Sign Language Translation}

Because the full task involving video to text translation was unapproachable at the time, early work on sign language translation focused on generation cascaded through \textit{glosses}, which are nonstandardized linguistic annotations representing signs. This allowed the task to be formulated as a special case of (sentence-level) text-to-text translation and reuse methods from mainstream MT~\citep{chapman1997lexicon,veale1998challenges,earlySlt}.

Unlike MT for written languages, translation from sign language glosses as a source representation is not immediately useful, because signers in general do not use them---only linguists and to some extent students do. Therefore the other half of the cascaded sign language understanding pipeline is sign language recognition (SLR), the task of predicting glosses from videos of people signing. Isolated SLR classifies a single gloss from a short clip~\citep{isolatedSLR,joze2019msasl,li2020word,desai2023asl,popsign}, and continuous SLR predicts a sequence of glosses from a clip of an entire sentence~\citep{koller15:cslr,Cui_2017_CVPR}. This sentence granularity is inherited from translation above and by analogy to automatic speech recognition (ASR), but is not especially harmful here: context is not strictly necessary because the task is to transcribe form, not understand meaning.

The modern framing for end-to-end video-to-text sign language MT originates in \citet{slt_orig}. The paper does not phrase the sentence-level framing as an explicit decision point but rather inherits it again from mainstream machine translation and continuous SLR. Because videos (and more generally, long sequences) are computationally difficult to work with/learn from, there is also an unstated pressure to use shorter clips. The provided dataset, RWTH-PHOENIX-Weather 2014T, is constructed on top of an existing (sentence-level) continuous SLR dataset~\citep{koller15:cslr} of weather forecasts interpreted into German Sign Language. There is no human baseline provided for the task, but if there were, it would likely be uneventful due to the dataset's limited domain and non-native interpreters.

As subsequent datasets have expanded into more sign languages and broader domains (and de-emphasized glosses, because they are a lossy bottleneck with limited availability), the datasets have retained the sentence-level framing---despite being constructed from long video corpora, where full discourse context is available and where there is not necessarily a sentence-level correspondence between the speech and sign tracks. Human annotations have been used to preprocess/quality check the dataset~\citep{content4all,bobsl,openasl,joshi2023isltranslate,shen2023auslandaily,uthus2023youtubeasl} or evaluate model outputs~\citep{wmt_slt,wmt_slt_23,how2sign}, but not to explore the sentence-level framing itself. See Appendix~\ref{sec:related-work-expanded} for a dataset-by-dataset analysis.

While surveying gloss-based translation methods,~\citet{muller2022considerations} note that only sentence-level systems had been studied at the time, and they give spatial indexing as one example of a grammatical feature that may be truncated in sentence-level systems. We are aware of only one work that has studied sign language translation beyond the sentence level since then,~\citet{sincan2023context}. Their work examines the empirical gains from providing models with prior text context---either full sentences or sign spottings---without specific sign linguistic motivation. Quality improves significantly but is still extremely low in absolute, so it is possible that the context is being used as a shortcut rather than an essential part of the task framing. Our work is complementary in that we analyze a wide variety of linguistic phenomena, and study a setting (human performance) where we are not bottlenecked by limitations of current training datasets and can more easily interpret results qualitatively.

\subsection{Document-Level Translation}

While the majority of work on machine translation focuses on (and has been very successful within) the sentence-level task framing, there is a body of work that highlights the aspects that are lost between sentences. Automatic reference-based metrics are relatively insensitive to discourse-level problems that stand out to human raters~\citep{hardmeier,laubli2018machine}, such as issues with lexical consistency, formality, and gender/number agreement~\citep{spokenContext,fernandes2023does}. Therefore many works create contrastive test sets where several candidate translations are ranked with respect to each other, rather than translations being generated from a blank slate, to measure these properties~\citep{bawden-etal-2018-evaluating,muller2019largescale,nagata-morishita-2020-test}. These works mostly evaluate model outputs rather than ideal (human) performance, but, e.g.,~\citet{matsuzaki-etal-2015-evaluating} provides a human baseline for English$\rightarrow$Japanese translation of short dialogues, in which the rate of correct translations is 18 percentage points higher given full document context vs. only an isolated sentence. We extend this line of work to sign languages by surveying extra linguistic phenomena related to the visual-spatial modality, then evaluate the empirical importance of discourse-level effects in this domain using a combination of automatic metrics and human ratings in the ideal (human) setting.

Historically, the bottleneck for training document-level MT has been the availability of document-level parallel corpora~\citep{spokenContext}; only a small fraction of translation data was natively document-level, such as video content with subtitles in multiple languages~\citep{lison-etal-2018-opensubtitles2018,duh18multitarget}.\footnote{Recently with the rise of self-supervised pretraining and LLMs this is less of a concern, since document-level monolingual data is abundant~\citep{siddhant-etal-2020-leveraging,wang-etal-2023-document-level}.} The situation is markedly different for sign languages: virtually all sign language corpora are natively discourse-level (with minor exceptions like SP-10~\cite{sp10} and WMT-SLT Signsuisse~\citep{wmt_slt_23}, which consist of isolated dictionary example sentences) but are preprocessed into isolated clips. Why not use this extra structure?

\begin{figure*}[t]
    \centering
    \includegraphics[scale=0.63]{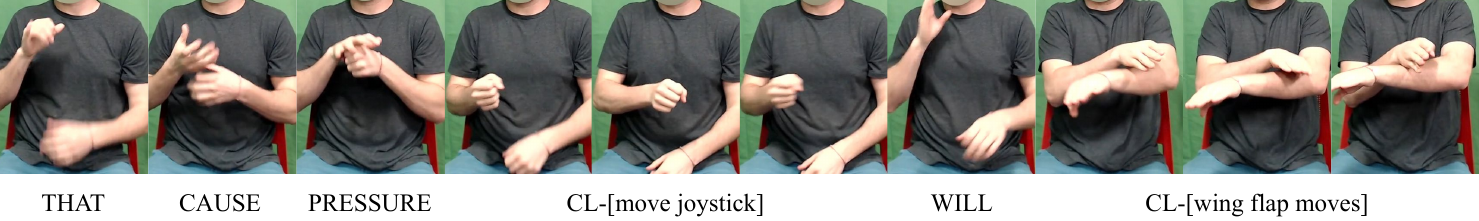}
    \vspace{-2mm}
    \caption{\textbf{Example of the interaction between classifiers and long-range context.} It isn't clear in isolation that the fist moving back and forth represents a fist controlling a joystick, or that the arm represents the plane's wing and the hand represents a flap (aileron) on the wing. Interpreter's head omitted here for privacy.}
    \label{fig:aileron}
\end{figure*}

\section{Long-Range Linguistic Dependencies}
\label{sec:linguistics}

In this section, we outline a number of long-range dependencies in the grammar of sign languages, primarily ASL, which may be truncated with sentence-level clipping. These features are not necessarily universal to all sign languages, but they are relatively common insofar as they are motivated by the visual-spatial modality~\citep{texas2002modality,universalMorphology}.

We create example figures using clips from the How2Sign dataset~\cite{how2sign}; we omit the signers' faces in the figures for privacy but note that facial expressions and mouthing are important in sign language.

\subsection{Spatial Referencing}

Perhaps the most salient feature that distinguishes sign languages from spoken languages is the ability to use space in a way that is grammatically structured (as opposed to in co-speech gesture)~\citep{useOfSpace}.

\subsubsection*{Pronouns}

Whereas spoken languages use third-person pronouns to refer to entities that were previously introduced in the discourse, sign languages use \textit{spatial indexing}, i.e., they establish that a locus in space refers to a particular entity and then reference that entity by pointing ~\citep{useOfSpace,invisibleSurrogates}. Because spoken languages tend to have a small set of third-person pronouns, they become ambiguous as the number of entities under discussion grows. But the number of unambiguous referents in sign languages may grow as space and memory permit, especially when using more complex forms of reference than pointing~\citep{referencingCrosslinguistically}.

So it may be the case that a spatial index in a sign language should be translated into a named entity in a spoken language (rather than a pronoun), or vice versa---but without context, it's impossible to know what name corresponds to that spatial index, or where that named entity lies in space. This is like a more severe version of translation between languages that have gendered vs. ungendered (or omissible) pronouns~\citep{frank2004gender,genderBias}.

\subsubsection*{Directional Verbs}

Some verbs in sign languages are \textit{directional}, i.e., their movement is inflected to agree with the spatial loci of their arguments~\citep{liddell1990four}. This is analogous to polypersonal agreement in spoken languages (verb agreement with respect to multiple arguments), but more flexible (and more context-dependent) for the same reason as pronouns above.

\subsubsection*{Classifiers}

In certain spoken languages, the term ``classifiers'' refers to words that agree with nouns of different semantic categories, and are often obligatory when counting nouns with numerals~\citep{classifiers}. In sign languages, classifiers are more expansive: like with spoken classifiers, different handshapes represent different categories of objects, but they can also be inflected in \textit{classifier predicates}, where the location and movement of the classifier take on an extremely flexible, iconic predicative meaning~\citep{aslClassifiers,Liddell1980}. A classic example is the 3 handshape in ASL (extended thumb, index, and middle finger) oriented with the thumb up, which represents a number of vehicles, especially cars. The classifier can be repeated across space to describe a packed parking lot, swerved side to side to depict a car driving down a winding road, slammed into another surface to represent a car crash, etc.

Because classifiers can refer to many objects in a particular category, and the referent needs only be clear from context (either explicitly introduced or just implied by the situation), the subject or entire meaning of a classifier predicate may not be clear in isolation. For example, in Figure~\ref{fig:aileron} it is only clear from context that the classifiers are referring to a joystick \& wing flaps in an airplane.

\subsubsection*{Role Shift}

When describing interactions between two or more characters, signers will often \textit{role shift}, i.e., they physically embody and act out the different characters~\citep{padden1986verbs}. This is analogous to quotes in spoken languages, except that turn-taking is not marked explicitly with words like ``he said'': instead, it's marked by shifting the body's angle/position and demeanor. In sentence-level clips, it may not be clear who is referenced by each role---or even that role shift is being used at all---because each turn in the role shift is considered its own sentence and clipped in isolation.

\subsection{Out-of-Vocabulary Terms}
\label{sec:oov-terms}

With languages in the same modality, it is usually straightforward to translate out-of-vocabulary terms like proper nouns by copying them directly from the source into the target context (perhaps with some phonological tweaks and transliteration, complicated somewhat by acronyms). But this strategy breaks down across modalities.

Because spoken languages are socially dominant over sign languages, virtually every sign language can productively borrow terms from spoken languages, through \textit{fingerspelling} (spelling the word with a manual alphabet) or \textit{mouthing} (silently saying the word while producing a related sign). But the reverse isn't true: spoken languages have no mechanism for borrowing signs. Context is important for strategies that reconcile this mismatch.

\subsubsection*{Abbreviated Fingerspelling}

\begin{figure}[t]
    \centering
    \includegraphics[scale=0.4]{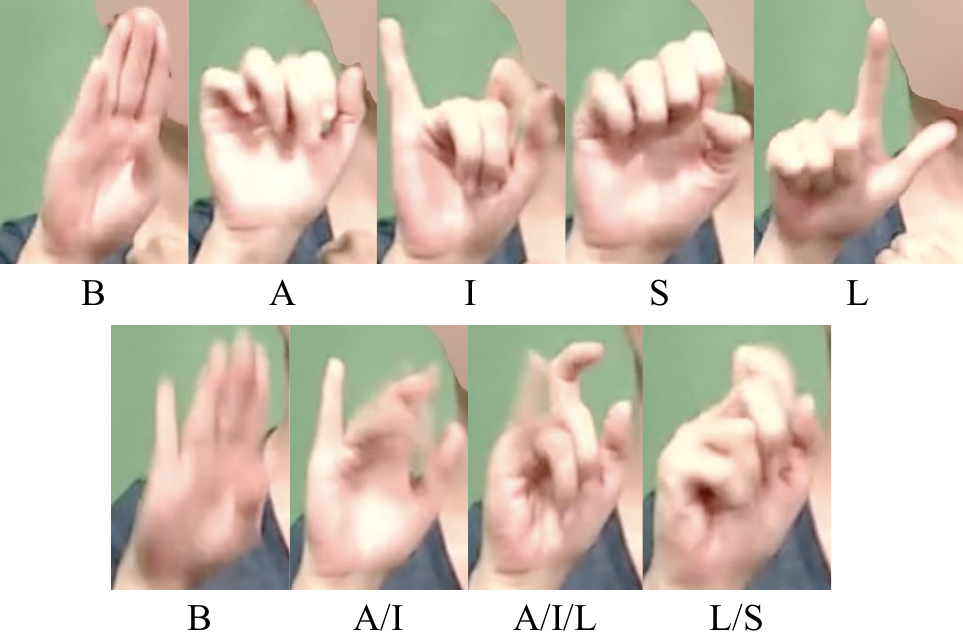}
    \vspace{-2mm}
    \caption{\textbf{Example of the interaction between rapid fingerspelling and long-range context.} Top is the first ``basil'' in the narrative (itself spelled slightly out of order), and bottom is the version from the test sentence: highly coarticulated, with multiple letters produced simultaneously. The labels indicate the relevant letters given the ground truth, but without context other letters such as Y, X, and T could be perceived.}
    \label{fig:basil}
    \vspace{-4mm}
\end{figure}

When introducing a fingerspelled term for the first time in a discourse, signers will spell it clearly to make sure that it can be understood. But when returning to that term later, they may speed through it amorphously to save time, with the understanding that the viewer can recognize the shape of the word in context. For example, in Figure~\ref{fig:basil} the letters of the word ``basil'' are fingerspelled simultaneously and out of order. This is described as ``careful'' vs. ``rapid'' fingerspelling in the literature~\citep{patrie,thumann,wager}.\footnote{A similar reduction happens for repeated spoken words too, but the effect is smaller~\citep{pmid26089592}.} 

If the signer anticipates that they will refer to the term repeatedly, especially for proper nouns, they may even declare a temporary acronym upfront and use it for the remainder of the discourse. For example, in an instance from the human baseline the trading card ``Whalebone Glider'' is abbreviated ``WG'' after its first mention. Absent context, it is difficult or impossible for someone viewing sentence-level clips to know what these abbreviated terms refer to, and copying the abbreviations directly would be unnatural in the target spoken language. The other translation direction is perhaps less problematic, because one could guess whether a proper noun is being used for the first time based on local cues and translate appropriately.

\subsubsection*{Name Signs}

In American Deaf culture, in addition to their full legal names, signers use sign names given to them by other members of the Deaf community. If their name is short enough, a person's sign name may be a fingerspelled version of their first name, but otherwise it is an idiosyncratic sign based on factors like their personality, appearance, and interests; name signs are perhaps even more idiosyncratic than names in spoken languages~\citep{namesigns}. When talking to an unfamiliar audience, a signer will often fingerspell a person's name and give their name sign, then refer to them using their name sign for the rest of the discourse. Training on isolated clips that include name signs will encourage the model to hallucinate. Challenges with name signs are not necessarily universal across sign languages; for example, in Japanese Sign Language, name signs are often a function of the kanji in a signer's legal name~\citep{jslNameSigns}, and therefore could more easily be translated without context.

\subsubsection*{Nonstandard signs}

For a variety of historical reasons---lack of a writing system, the very recent development of video calling, historical exclusion of sign languages from education---ASL lacks standardized vocabulary in certain academic fields~\citep{mckee}.\footnote{There are \href{https://aslcore.org/}{efforts} underway to invent standardized vocabulary, but currently each school or even each class tends to invent its own signs as needed.} When introducing a nonstandard or niche sign, the signer will often fingerspell it to ensure that it is understood by a less familiar audience.
When translating from a sign language into a spoken language, like with name signs the model may be able to guess the meaning but is generally encouraged to hallucinate. When translating from a spoken language into a sign language, if the model knows multiple nonstandard signs it is unclear how it could coordinate their usage across independently translated sentences, as seen with lexical cohesion issues in MT for written languages~\citep{spokenContext}.

\subsection{Generic Context Dependence}

In addition to the aforementioned features specific to sign languages and the visual-spatial modality, sign languages can be context-dependent in similar ways to spoken languages. For example, in terms of grammar: ASL can drop pronouns~\citep{prodrop} and has a variety of strategies for expressing tense~\citep{tense} and definiteness/indefiniteness~\citep{definiteness}. In terms of vocabulary: lexical signs can be ambiguous or dialectal, making them harder to understand without context.

\section{Case Study}
\label{sec:evaluation}

In order to explore how these phenomena surface in real sign language translation datasets, we perform a human baseline for ASL to English translation on How2Sign \cite{how2sign} across different amounts of provided context. To the best of our knowledge, this is the first time human performance has been measured for the sentence-level sign language machine translation task.

How2Sign was constructed by having 11 signers---5 Hearing, 4 Deaf, and 2 Hard of Hearing---watch English-captioned instructional ``how to'' videos from the earlier How2 dataset~\citep{how2} a first time to understand the content, then a second time at 0.75x speed while performing a live interpretation. The captions (from the original speech track) were manually realigned to the signing, with an average sentence duration of 8.67 seconds.

\subsection{Setup}
\label{sec:setup}

First, we describe the human baseline test instances and settings. Here in the context of ASL to English translation, we use \textbf{s} to refer to the source ASL clips for a particular video and \textbf{t} to refer to its target English captions. $i$ is the index of a particular clip/caption within that video. We collect translations across four different context settings:

\vspace{-2mm}\begin{itemize}
  \setlength\itemsep{-1mm}
    \item \textbf{s$_i$}: The source clip alone. This is the classic sign language machine translation framing.
    \item \textbf{s$_{i-1:i}$}: The source clip extended backwards to include the previous clip.
    \item \textbf{s$_{i-1:i}$}, \textbf{t$_{i-1}$}: The previous and current source clip, plus the ground truth text for the previous clip.\footnote{Using the ground truth is slightly unrepresentative of what is possible at test time; the ideal would have been to translate using the entire source video up to this point as context, but evaluating this setting would have been prohibitively time-consuming. These settings that condition on previous captions are more similar to how we expect machine learning practitioners to incorporate context in light of sequence length constraints initially, like in~\citet{sincan2023context}.}
    \item \textbf{s$_{i-1: i}$}, \textbf{t$_{0:i-1}$}: The previous and current source clip, plus the ground truth captions for the entire video up to this point.
\end{itemize}\vspace{-2mm}

Note that each of these settings strictly expands upon the prior one, so it is valid for a single annotator to perform all four in sequence. (Some of these translations may be identical to those for prior settings, if the annotator does not want to adjust their translation in light of new context.) However, it is not valid for an annotator to translate multiple clips $i$ within a single video due to leakage. On top of these four translation settings, we also ask the annotators to describe how well they understood the sentence in isolation vs. after seeing additional context, and to rate the naturalness of the signing on a scale from 0-2, where higher is more natural.\footnote{Specifically, they were asked to answer ``Is it natural ASL?'', with 0=``no'', 1=``eh'', and 2=``yes'' as the options.}

To select our human baseline instances we start with How2Sign's test set, which consists of 184 ASL translations of 149 How2 narratives, sliced into 2,322 clips. We discard narratives that are translated multiple times by different signers (to avoid cross-instance leakage) and videos that seem generally malformed (e.g., large spans of the video lack captions or captions extend beyond the duration of the video). For each remaining narrative, we sample a clip uniformly at random, excluding the first clip in each narrative because results for the context settings would be trivial.\footnote{This means that our metrics will slightly overestimate the effect of context, because they ignore initial sentences that are meant to be understood without context.} Some clips within narratives are not contiguous because the signer made an error between sentences, which breaks the \textbf{s$^{i-1:i}$} condition; we reject these cases and resample until success. The result is a set of 102 test instances, at most one per narrative.

Second, we describe the actual execution of the human baseline: Our annotators were the two middle authors, who are Deaf signers who use both ASL and English as primary languages;\footnote{Note that these annotators are not professional translators, which may harm the quality of the translated outputs (and automated metrics computed on them). However, the English captions in How2Sign (originally from How2) are not especially polished themselves, since they are transcriptions of spontaneous speech with disfluencies etc., so we expect this to be less of an issue than if we were comparing to reference translations by professional sign language interpreters of originally signed content. These annotators also know the research purpose (and could have inferred it from the sequence of context settings, even if they hadn't had foreknowledge), which may bias the translations and ratings. We were more concerned with getting a good qualitative understanding of the data amongst the authors.} the other authors set up the test instances. Each annotator spent several hours performing the translations and ratings for a random nonoverlapping split of the data, leaving additional commentary as they went for use in our qualitative analysis. The annotators were allowed to slow down or repeat the video, but were told not to agonize over it frame by frame. See Appendix~\ref{app:human-baseline-instructions} for annotator instructions.

\subsection{Results}
\label{sec:results}

Following prior works that evaluate on How2Sign~\citep{alvarezsign, lin2023glossfree, tarres2023sign, uthus2023youtubeasl}, we report BLEU~\citep{papineni-etal-2002-bleu} and BLEURT~\citep{sellam2020bleurt} as our quantitative metrics. We compute BLEU using SacreBLEU \cite{post-2018-call} version 2 with all default options, and BLEURT using checkpoint BLEURT-20~\cite{bleurt20paper}. See Table~\ref{tab:scores} for scores, Table~\ref{tab:ratings} for ratings, and Appendix~\ref{app:human-baseline-results} for the complete set of translations comprising the human baseline.

\begin{table}[t]
    \centering
    \small
    \setlength{\tabcolsep}{4pt}
    \begin{tabular}{lcccc}
    \toprule
    \bf BLEU& \textbf{s$_i$} & \textbf{s$_{i-1:i}$} & \textbf{s$_{i-1:i}$}, \textbf{t$_{i-1}$} & \textbf{s$_{i-1: i}$}, \textbf{t$_{0:i-1}$} \\
    \midrule
    Average & 19.8 & 20.4 & 21.1 & \underline{21.5} \\
    \midrule
    Interpreter A & 5.2 & 6.0 & 6.1 & \underline{6.3} \\
    Interpreter B & 18.4 & 19.1 & 20.5 & \underline{21.0} \\
    Interpreter C & 7.4 & 7.2 & 8.2 & \underline{8.7} \\
    Interpreter D & 39.5 & 40.9 & \underline{41.3} & 41.1 \\
    Interpreter E & 19.4 & 19.5 & 19.8 & \underline{20.7} \\
    \bottomrule
    \end{tabular} \\
    \vspace{2mm}
    \begin{tabular}{lcccc}
    \toprule
    \bf BLEURT & \textbf{s$_i$} & \textbf{s$_{i-1:i}$} & \textbf{s$_{i-1:i}$}, \textbf{t$_{i-1}$} & \textbf{s$_{i-1: i}$}, \textbf{t$_{0:i-1}$} \\
    \midrule
    Average & 56.6 & 58.1 & 59.4 & \underline{59.5} \\
    \midrule
    Interpreter A & 45.7 & 48.7 & \underline{49.3} & 48.6 \\
    Interpreter B & 57.3 & 57.1 & 57.6 & \underline{58.1} \\
    Interpreter C & 47.7 & 50.0 & 54.3 & \underline{55.0} \\
    Interpreter D & 70.0 & 70.6 & \underline{71.1} & 70.3 \\
    Interpreter E & 59.7 & 61.3 & 61.3 & \underline{61.8} \\
    \bottomrule
    \end{tabular}
    \caption{BLEU (top) and BLEURT (bottom) scores ($\uparrow$) for the human baseline for ASL to English translation, across different amounts of provided context and different interpreters featured in the videos.}
    \label{tab:scores}
    \vspace{-2mm}
\end{table}

\paragraph*{Effect of context.} Human performance on the sentence-level translation task is 19.8 BLEU (56.6 BLEURT) and increases monotonically with extra context, but only up to 21.5 BLEU (59.5 BLEURT). This consistent but relatively small difference in automatic metrics belies the annotators' perception of the gap: for 33.3\% of test instances, the annotators judged that they were unable to understand key details of the signed content from the sentence in isolation which they later understood from additional context (verified with their actual translations across settings compared to the ground truth). Of these failure cases, 47\% featured classifiers with unclear referents, 38\% grammatical features like prodrop/lack of overt tense markings, 26\% rapid fingerspelling, 9\% acronyms, 6\% ambiguous signs, and 6\% dialectal sign variation.\footnote{We didn't come across any How2Sign instances of several linguistic phenomena described in Section~\ref{sec:linguistics}, for a variety of presumed reasons. Spatial indexing, directional verbs, and role shift are relevant when discussing third-person characters (especially multiple ones interacting), but How2Sign is largely first-person or second-person given the instructional narrative domain. Name signs are generally only used in originally produced Deaf content. Nonstandard signs are used primarily by domain experts, so they are unlikely to be introduced in content translated from English without much preparation.}

In addition to translations that improved given past context, there were several examples where the translation was incorrect across all settings because future context was needed to understand the sentence. We did not anticipate this, so there was no experimental setting to measure it.

\paragraph*{Variation across interpreters.}
We observe qualitatively that there is enormous variation in signing style between the five interpreters (which we label A-E) featured in the test videos, across the spectrum from ASL to CASE to MCE. It is hard to disentangle this from the shallow translations that are typical of live interpreting.
Inspired by prior work which disaggregates evals~\citep{pmlr-v81-buolamwini18a, auditing, barocas2021designing,kaplun2022deconstructing}, we therefore break down our results by interpreter.

We find that the human baseline metrics match our subjective impressions: they vary from 5.2 BLEU (45.7 BLEURT) for Interpreter A to 39.5 (70.0) for Interpreter D. The interpreters with lower scores perform deeper translation closer to ASL, and those with higher scores border on MCE (which inflates n-gram overlap, because the task approaches sign recognition rather than translation). The interpreters signing with more English influence also tend to mouth more prominently, so sometimes the translation is clear from lipreading even when the signing itself is odd and hard to understand. The annotators rated the average naturalness of the content at 1.05 on a scale from 0-2 ($\uparrow$), ranging from 1.93 for Interpreter A to 0.64 for Interpreter D; generally, the more natural the content, the worse the sentence-level translation metrics.\footnote{We emphasize that these naturalness judgments are subjective from the perspective of the annotators. This may be biased by social factors like the perception that a hypercorrect ``pure'' form of ASL is the most prestigious, as opposed to signing with more influence from English---or vice versa~\citep{stokoe1969sign,vicars2023alternating}. Sign language translation models should still understand this content (especially to the extent that this reflects real variation in how people sign, as opposed to performance effects of live interpreting), but it is important to know what we are actually evaluating so that we do not e.g. test on artificially easy content and overstate performance for actual Deaf signers.}

When we look at the other three settings, we see that context has a proportionally larger effect for interpreters where the translation metrics were originally lower (and naturalness is rated higher): Interpreter A increases from 5.2 BLEU (45.7 BLEURT) to 6.3 (48.6) and Interpreter C from 7.4 (47.7) to 8.7 (55.0), vs. Interpreter D from 39.5 (70.0) to 41.1 (70.3). This bears out in the annotator ratings as well: translation failed due to missing context 73.3\% of the time on Interpreter A and 44.0\% of the time on Interpreter C, but only 13.6\% of the time on Interpreter D. This confirms our suspicion that the effect of discourse context is obscured by evaluating on live (and especially hearing) interpreters. Even though there is a clear improvement in metrics due to context, the average effect size is obscured by the fact that we are essentially evaluating on multiple domains at once.

\paragraph*{Misalignment.} Despite How2Sign's use of manually realigned captions (and despite us having excluded apparently malformed videos earlier), 5\% of the sentence-level clips in our baseline still do not contain the relevant content. Even more clips lack significant parts of the ground truth translation or have extra content beyond it. On top of this, the onset of a sentence usually begins earlier on the face than the hands, so with even with ``accurate'' clipping the sentence may either start with a leftover handshape from the previous sentence or truncate the start of the sentence on the face. These all combine to make it difficult for annotators (or models) to know which parts of the input clip they should and shouldn't translate. In a discourse-level framing, misalignment matters less because the offset is a smaller fraction of the overall content---or there is no misalignment at all if the entire discourse is in the translation context.

\begin{table}[t]
    \centering
    \small
    \begin{tabular}{lcc}
    \toprule
    & \textbf{\% ctxt failure} & \textbf{naturalness (0-2, $\uparrow$)} \\
    \midrule
    Average & 33.3 & 1.05 \\
    \midrule
    Interpreter A & 73.3 & 1.93\\
    Interpreter B & 14.3 & 1.00 \\
    Interpreter C & 44.0 & 1.24\\
    Interpreter D & 13.6 & 0.64\\
    Interpreter E & 26.9 & 0.73 \\
    \bottomrule
    \end{tabular}
    \caption{\textbf{Annotator ratings for the human baseline}---\% of instances where they failed to understand key details from the sentence in isolation but later succeeded with context, and naturalness of the signed content on a scale from 0-2 ($\uparrow$)---broken down by interpreter.}
    \label{tab:ratings}
    \vspace{-2mm}
\end{table}

\section{Conclusion}

In this paper, we argued that the costs of the sentence-level sign language MT task framing are higher than many might expect from experience with spoken languages, with many relevant discourse-level phenomena being related to the visual-spatial modality and cross-modal translation. We supported this with a case study: the first human baseline for sentence-level sign language MT, from ASL to English on the How2Sign dataset. We found that discourse context was necessary to fully understand and translate a large fraction of sentences (33.3\%), and this effect is itself attenuated by the prevalence of signing data that does not represent the more challenging aspects of ASL due to its use of non-native or live interpreters. We hope that this inspires more in-depth analysis grounded in firsthand experience with sign languages, to avoid perpetuating systemic bias in the way we conceptualize sign language tasks~\citep{desai2024systemic}.

\section*{Limitations}
Our results are limited in that we empirically evaluate one language pair (ASL and English), one translation direction (ASL to English), and one domain (instructional narratives from the \emph{How2Sign} dataset). Extrapolating from our analysis in Section~\ref{sec:linguistics}:
\begin{itemize}
    \item We expect the aforementioned long-range dependencies to exist in other sign languages, because they are generally motivated by features of the visual-spatial modality.
    \item We expect English to ASL translation (translation from a spoken language into a signed language) to suffer similar problems. Sometimes, source sentences would not include enough grounding to perform a natural translation with classifiers. And even when source sentences do include all necessary information to perform a faithful translation, even a perfect sentence-level translation model would result in unnatural discourse-level translations when concatenating clips due to inconsistent use of space and other discourse phenomena across sentences.
    \item Direct translation between two sign languages may be less problematic than translation between a sign language and a spoken language, because similarities in use of space or classifiers may allow for a shallower translation.
    \item Results from \emph{How2Sign} may not be representative of results on other domains. Informal instructional narratives are relatively well-suited to showing the inadequacies of sentence-level translation, because they are grounded in a single scene for the duration of the narrative and use relatively short sentences. However, they are also light on description of multiple third-person entities interacting with each other, which use other context-dependent structures described above. We expect stories/ASL literature to require more context, and content with stronger influence from English (or the respective dominant spoken language for other regions) to require less.
\end{itemize}

\section*{Ethics Statement}

The ethical considerations of this work are those for sign language processing as a whole. Namely, machine understanding of sign languages would improve access to information, communication, and other technologies for underserved signing communities. However, there is a risk that---rather than supplement existing resources to strictly improve access---entities who currently provide services in sign languages might replace a high-quality solution that uses human interpreters with a lower-quality automated one. This work tries to expose deficits in the current task framing so that automatic solutions will be less flawed. Inclusion in modern NLP also brings with it a number of well-known risks (misinformation, bias, etc. at scale). Future works that release trained models should mitigate these potential harms.

\section*{Acknowledgements}
We thank Chris Dyer and Manfred Georg for giving feedback on drafts of this paper and Caroline Pantofaru for institutional support.

\bibliography{custom}

\begin{thebibliography}{83}
\providecommand{\natexlab}[1]{#1}

\bibitem[{Albanie et~al.(2021)Albanie, Varol, Momeni, Bull, Afouras, Chowdhury,
  Fox, Woll, Cooper, McParland, and Zisserman}]{bobsl}
Samuel Albanie, Gül Varol, Liliane Momeni, Hannah Bull, Triantafyllos Afouras,
  Himel Chowdhury, Neil Fox, Bencie Woll, Rob Cooper, Andrew McParland, and
  Andrew Zisserman. 2021.
\newblock \href {https://doi.org/10.48550/ARXIV.2111.03635} {Bbc-oxford british
  sign language dataset}.
\newblock \emph{arXiv preprint}.

\bibitem[{Allan(1977)}]{classifiers}
Keith Allan. 1977.
\newblock \href {https://doi.org/10.1353/lan.1977.0043} {Classifiers}.
\newblock \emph{Language}, 53:285--311.

\bibitem[{{\'A}lvarez et~al.(2022){\'A}lvarez, Nieto, and Benet}]{alvarezsign}
Patricia~Cabot {\'A}lvarez, Xavier~Gir{\'o} Nieto, and Laia~Tarr{\'e}s Benet.
  2022.
\newblock \href
  {https://imatge.upc.edu/web/publications/sign-language-translation-based-transformers-how2sign-dataset}
  {Sign language translation based on transformers for the {How2Sign} dataset}.

\bibitem[{Aronoff et~al.(2005)Aronoff, Meir, and Sandler}]{universalMorphology}
Mark Aronoff, Irit Meir, and Wendy Sandler. 2005.
\newblock \href {https://doi.org/10.1353/lan.2005.0043} {The paradox of sign
  language morphology}.
\newblock \emph{Language}, 81(2):301–344.

\bibitem[{B.~Shi and Livescu(2019)}]{fs18iccv}
J.~Keane D. Brentari G.~Shakhnarovich B.~Shi, A. Martinez Del~Rio and
  K.~Livescu. 2019.
\newblock Fingerspelling recognition in the wild with iterative visual
  attention.
\newblock \emph{ICCV}.

\bibitem[{B.~Shi and Livescu(2018)}]{fs18slt}
J.~Keane J. Michaux D. Brentari G.~Shakhnarovich B.~Shi, A. Martinez Del~Rio
  and K.~Livescu. 2018.
\newblock American sign language fingerspelling recognition in the wild.
\newblock \emph{SLT}.

\bibitem[{Barocas et~al.(2021)Barocas, Guo, Kamar, Krones, Morris, Vaughan,
  Wadsworth, and Wallach}]{barocas2021designing}
Solon Barocas, Anhong Guo, Ece Kamar, Jacquelyn Krones, Meredith~Ringel Morris,
  Jennifer~Wortman Vaughan, Duncan Wadsworth, and Hanna Wallach. 2021.
\newblock \href {https://arxiv.org/abs/2103.06076} {Designing disaggregated
  evaluations of ai systems: Choices, considerations, and tradeoffs}.
\newblock \emph{Preprint}, arXiv:2103.06076.

\bibitem[{Bawden et~al.(2018)Bawden, Sennrich, Birch, and
  Haddow}]{bawden-etal-2018-evaluating}
Rachel Bawden, Rico Sennrich, Alexandra Birch, and Barry Haddow. 2018.
\newblock \href {https://doi.org/10.18653/v1/N18-1118} {Evaluating discourse
  phenomena in neural machine translation}.
\newblock In \emph{Proceedings of the 2018 Conference of the North {A}merican
  Chapter of the Association for Computational Linguistics: Human Language
  Technologies, Volume 1 (Long Papers)}, pages 1304--1313, New Orleans,
  Louisiana. Association for Computational Linguistics.

\bibitem[{Bragg et~al.(2019)Bragg, Koller, Bellard, Berke, Boudreault,
  Braffort, Caselli, Huenerfauth, Kacorri, Verhoef, Vogler, and
  Ringel~Morris}]{bragg}
Danielle Bragg, Oscar Koller, Mary Bellard, Larwan Berke, Patrick Boudreault,
  Annelies Braffort, Naomi Caselli, Matt Huenerfauth, Hernisa Kacorri, Tessa
  Verhoef, Christian Vogler, and Meredith Ringel~Morris. 2019.
\newblock \href {https://doi.org/10.1145/3308561.3353774} {Sign language
  recognition, generation, and translation: An interdisciplinary perspective}.
\newblock In \emph{Proceedings of the 21st International ACM SIGACCESS
  Conference on Computers and Accessibility}, ASSETS '19, page 16–31, New
  York, NY, USA. Association for Computing Machinery.

\bibitem[{Buolamwini and Gebru(2018)}]{pmlr-v81-buolamwini18a}
Joy Buolamwini and Timnit Gebru. 2018.
\newblock \href {https://proceedings.mlr.press/v81/buolamwini18a.html} {Gender
  shades: Intersectional accuracy disparities in commercial gender
  classification}.
\newblock In \emph{Proceedings of the 1st Conference on Fairness,
  Accountability and Transparency}, volume~81 of \emph{Proceedings of Machine
  Learning Research}, pages 77--91. PMLR.

\bibitem[{Camgoz et~al.(2018)Camgoz, Hadfield, Koller, Ney, and
  Bowden}]{slt_orig}
Necati~Cihan Camgoz, Simon Hadfield, Oscar Koller, Hermann Ney, and Richard
  Bowden. 2018.
\newblock Neural sign language translation.
\newblock In \emph{Proceedings of the IEEE Conference on Computer Vision and
  Pattern Recognition (CVPR)}.

\bibitem[{Camgoz et~al.(2021)Camgoz, Saunders, Rochette, Giovanelli, Inches,
  Nachtrab-Ribback, and Bowden}]{content4all}
Necati~Cihan Camgoz, Ben Saunders, Guillaume Rochette, Marco Giovanelli,
  Giacomo Inches, Robin Nachtrab-Ribback, and Richard Bowden. 2021.
\newblock \href {https://doi.org/10.48550/ARXIV.2105.02351} {Content4all open
  research sign language translation datasets}.
\newblock \emph{arXiv preprint}.

\bibitem[{Chapman(1997)}]{chapman1997lexicon}
Robbin~Nicole Chapman. 1997.
\newblock \emph{A lexicon for translation of American Sign Language to
  English}.
\newblock Ph.D. thesis, Massachusetts Institute of Technology.

\bibitem[{Charayaphan and Marble(1992)}]{isolatedSLR}
C.~Charayaphan and A.~E. Marble. 1992.
\newblock {{I}mage processing system for interpreting motion in {A}merican
  {S}ign {L}anguage}.
\newblock \emph{J Biomed Eng}, 14(5):419--425.

\bibitem[{Coster et~al.(2023)Coster, Shterionov, Herreweghe, and
  Dambre}]{decosterSurvey}
Mathieu~De Coster, Dimitar Shterionov, Mieke~Van Herreweghe, and Joni Dambre.
  2023.
\newblock \href {https://doi.org/10.1007/s10209-023-00992-1} {Machine
  translation from signed to spoken languages: state of the art and
  challenges}.
\newblock \emph{Universal Access in the Information Society}.

\bibitem[{Cui et~al.(2017)Cui, Liu, and Zhang}]{Cui_2017_CVPR}
Runpeng Cui, Hu~Liu, and Changshui Zhang. 2017.
\newblock Recurrent convolutional neural networks for continuous sign language
  recognition by staged optimization.
\newblock In \emph{Proceedings of the IEEE Conference on Computer Vision and
  Pattern Recognition (CVPR)}.

\bibitem[{Desai et~al.(2023)Desai, Berger, Minakov, Milan, Singh, Pumphrey,
  Ladner, au2, Lu, Caselli, and Bragg}]{desai2023asl}
Aashaka Desai, Lauren Berger, Fyodor~O. Minakov, Vanessa Milan, Chinmay Singh,
  Kriston Pumphrey, Richard~E. Ladner, Hal Daumé~III au2, Alex~X. Lu, Naomi
  Caselli, and Danielle Bragg. 2023.
\newblock \href {https://arxiv.org/abs/2304.05934} {Asl citizen: A
  community-sourced dataset for advancing isolated sign language recognition}.
\newblock \emph{Preprint}, arXiv:2304.05934.

\bibitem[{Desai et~al.(2024)Desai, Meulder, Hochgesang, Kocab, and
  Lu}]{desai2024systemic}
Aashaka Desai, Maartje~De Meulder, Julie~A. Hochgesang, Annemarie Kocab, and
  Alex~X. Lu. 2024.
\newblock \href {https://arxiv.org/abs/2403.02563} {Systemic biases in sign
  language ai research: A deaf-led call to reevaluate research agendas}.
\newblock \emph{Preprint}, arXiv:2403.02563.

\bibitem[{Duarte et~al.(2021)Duarte, Palaskar, Ventura, Ghadiyaram, DeHaan,
  Metze, Torres, and Giro-i Nieto}]{how2sign}
Amanda Duarte, Shruti Palaskar, Lucas Ventura, Deepti Ghadiyaram, Kenneth
  DeHaan, Florian Metze, Jordi Torres, and Xavier Giro-i Nieto. 2021.
\newblock {How2Sign: A Large-scale Multimodal Dataset for Continuous American
  Sign Language}.
\newblock In \emph{Conference on Computer Vision and Pattern Recognition
  (CVPR)}.

\bibitem[{Duh(2018)}]{duh18multitarget}
Kevin Duh. 2018.
\newblock The multitarget ted talks task.
\newblock \url{http://www.cs.jhu.edu/~kevinduh/a/multitarget-tedtalks/}.

\bibitem[{Emmorey(1996)}]{useOfSpace}
Karen Emmorey. 1996.
\newblock \href {https://doi.org/10.7551/mitpress/4107.003.0007} {{The
  Confluence of Space and Language in Signed Languages}}.
\newblock In \emph{{Language and Space}}. The MIT Press.

\bibitem[{Fernandes et~al.(2023)Fernandes, Yin, Liu, Martins, and
  Neubig}]{fernandes2023does}
Patrick Fernandes, Kayo Yin, Emmy Liu, André F.~T. Martins, and Graham Neubig.
  2023.
\newblock \href {https://arxiv.org/abs/2109.07446} {When does translation
  require context? a data-driven, multilingual exploration}.
\newblock \emph{Preprint}, arXiv:2109.07446.

\bibitem[{Ferrara et~al.(2023)Ferrara, Anible, Hodge, Jantunen, Leeson, Mesch,
  and Nilsson}]{referencingCrosslinguistically}
Lindsay Ferrara, Benjamin Anible, Gabrielle Hodge, Tommi Jantunen, Lorraine
  Leeson, Johanna Mesch, and Anna-Lena Nilsson. 2023.
\newblock \href {https://doi.org/doi:10.1515/lingty-2021-0057} {A
  cross-linguistic comparison of reference across five signed languages}.
\newblock \emph{Linguistic Typology}, 27(3):591--627.

\bibitem[{Frank et~al.(2004)Frank, Hoffmann, Strobel et~al.}]{frank2004gender}
Anke Frank, Chr Hoffmann, Maria Strobel, et~al. 2004.
\newblock Gender issues in machine translation.
\newblock \emph{Univ. Bremen}.

\bibitem[{Frishberg(1975)}]{aslClassifiers}
Nancy Frishberg. 1975.
\newblock \href {http://www.jstor.org/stable/412894} {Arbitrariness and
  iconicity: Historical change in american sign language}.
\newblock \emph{Language}, 51(3):696--719.

\bibitem[{Gueuwou et~al.(2023)Gueuwou, Siake, Leong, and
  Müller}]{gueuwou2023jwsign}
Shester Gueuwou, Sophie Siake, Colin Leong, and Mathias Müller. 2023.
\newblock \href {https://arxiv.org/abs/2311.10174} {Jwsign: A highly
  multilingual corpus of bible translations for more diversity in sign language
  processing}.
\newblock \emph{Preprint}, arXiv:2311.10174.

\bibitem[{Hardmeier(2012)}]{hardmeier}
Christian Hardmeier. 2012.
\newblock \href {https://doi.org/10.4000/discours.8726} {Discourse in
  statistical machine translation: A survey and a case study}.
\newblock \emph{Discours}.

\bibitem[{Irani(2019)}]{definiteness}
Ava Irani. 2019.
\newblock \href {https://doi.org/10.5281/zenodo.3252018} {Chapter 4: On
  (in)definite expressions in american sign language.}

\bibitem[{Jacobowitz and Stokoe(1988)}]{tense}
E.~Lynn Jacobowitz and William~C. Stokoe. 1988.
\newblock \href {http://www.jstor.org/stable/26203876} {Signs of tense in asl
  verbs}.
\newblock \emph{Sign Language Studies}, (60):331--340.

\bibitem[{Jacobs et~al.(2015)Jacobs, Yiu, Watson, and Dell}]{pmid26089592}
C.~L. Jacobs, L.~K. Yiu, D.~G. Watson, and G.~S. Dell. 2015.
\newblock {{W}hy are repeated words produced with reduced durations? {E}vidence
  from inner speech and homophone production}.
\newblock \emph{J Mem Lang}, 84:37--48.

\bibitem[{Joshi et~al.(2023)Joshi, Agrawal, and Modi}]{joshi2023isltranslate}
Abhinav Joshi, Susmit Agrawal, and Ashutosh Modi. 2023.
\newblock \href {https://arxiv.org/abs/2307.05440} {Isltranslate: Dataset for
  translating indian sign language}.
\newblock \emph{Preprint}, arXiv:2307.05440.

\bibitem[{Joze and Koller(2019)}]{joze2019msasl}
Hamid Reza~Vaezi Joze and Oscar Koller. 2019.
\newblock \href {https://arxiv.org/abs/1812.01053} {Ms-asl: A large-scale data
  set and benchmark for understanding american sign language}.
\newblock \emph{Preprint}, arXiv:1812.01053.

\bibitem[{Kaplun et~al.(2022)Kaplun, Ghosh, Garg, Barak, and
  Nakkiran}]{kaplun2022deconstructing}
Gal Kaplun, Nikhil Ghosh, Saurabh Garg, Boaz Barak, and Preetum Nakkiran. 2022.
\newblock \href {https://arxiv.org/abs/2202.09931} {Deconstructing
  distributions: A pointwise framework of learning}.
\newblock \emph{Preprint}, arXiv:2202.09931.

\bibitem[{Koller et~al.(2015)Koller, Forster, and Ney}]{koller15:cslr}
Oscar Koller, Jens Forster, and Hermann Ney. 2015.
\newblock Continuous sign language recognition: Towards large vocabulary
  statistical recognition systems handling multiple signers.
\newblock \emph{Computer Vision and Image Understanding}, 141:108--125.

\bibitem[{Li et~al.(2020)Li, Rodriguez, Yu, and Li}]{li2020word}
Dongxu Li, Cristian Rodriguez, Xin Yu, and Hongdong Li. 2020.
\newblock Word-level deep sign language recognition from video: A new
  large-scale dataset and methods comparison.
\newblock In \emph{The IEEE Winter Conference on Applications of Computer
  Vision}, pages 1459--1469.

\bibitem[{Liddell(1990)}]{liddell1990four}
Scott Liddell. 1990.
\newblock Four functions of a locus: Reexamining the structure of space in asl.
\newblock In Ceil Lucas, editor, \emph{Sign Language Research: Theoretical
  Issues}, pages 176--198. Gallaudet University Press, Washington D.C.

\bibitem[{Liddell(2003)}]{invisibleSurrogates}
Scott Liddell. 2003.
\newblock \href {https://doi.org/10.1017/CBO9780511615054} {Grammar, gesture,
  and meaning in american sign language}.
\newblock \emph{Grammar, Gesture, and Meaning in American Sign Language}.

\bibitem[{Liddell(1980)}]{Liddell1980}
Scott~K. Liddell. 1980.
\newblock \href {https://doi.org/doi:10.1515/9783112418260} {\emph{American
  Sign Language Syntax}}.
\newblock De Gruyter Mouton, Berlin, Boston.

\bibitem[{Lillo-Martin(1986)}]{prodrop}
Diane Lillo-Martin. 1986.
\newblock \href {http://www.jstor.org/stable/4047639} {Two kinds of null
  arguments in american sign language}.

\bibitem[{Lin et~al.(2023)Lin, Wang, Zhu, Sun, Zhang, and
  Yang}]{lin2023glossfree}
Kezhou Lin, Xiaohan Wang, Linchao Zhu, Ke~Sun, Bang Zhang, and Yi~Yang. 2023.
\newblock \href {https://arxiv.org/abs/2305.12876} {Gloss-free end-to-end sign
  language translation}.
\newblock \emph{Preprint}, arXiv:2305.12876.

\bibitem[{Lison et~al.(2018)Lison, Tiedemann, and
  Kouylekov}]{lison-etal-2018-opensubtitles2018}
Pierre Lison, J{\"o}rg Tiedemann, and Milen Kouylekov. 2018.
\newblock \href {https://aclanthology.org/L18-1275} {{O}pen{S}ubtitles2018:
  Statistical rescoring of sentence alignments in large, noisy parallel
  corpora}.
\newblock In \emph{Proceedings of the Eleventh International Conference on
  Language Resources and Evaluation ({LREC} 2018)}, Miyazaki, Japan. European
  Language Resources Association (ELRA).

\bibitem[{Läubli et~al.(2018)Läubli, Sennrich, and Volk}]{laubli2018machine}
Samuel Läubli, Rico Sennrich, and Martin Volk. 2018.
\newblock \href {https://arxiv.org/abs/1808.07048} {Has machine translation
  achieved human parity? a case for document-level evaluation}.
\newblock \emph{Preprint}, arXiv:1808.07048.

\bibitem[{Matsuzaki et~al.(2015)Matsuzaki, Fujita, Todo, and
  Arai}]{matsuzaki-etal-2015-evaluating}
Takuya Matsuzaki, Akira Fujita, Naoya Todo, and Noriko~H. Arai. 2015.
\newblock \href {https://doi.org/10.3115/v1/P15-2024} {Evaluating machine
  translation systems with second language proficiency tests}.
\newblock In \emph{Proceedings of the 53rd Annual Meeting of the Association
  for Computational Linguistics and the 7th International Joint Conference on
  Natural Language Processing (Volume 2: Short Papers)}, pages 145--149,
  Beijing, China. Association for Computational Linguistics.

\bibitem[{McCaskill et~al.(2011)McCaskill, Lucas, Bayley, and Hill}]{blackasl}
Carolyn McCaskill, Ceil Lucas, Robert Bayley, and Joseph~Christopher Hill.
  2011.
\newblock \emph{The Hidden Treasure of Black ASL: Its History and Structure}.
\newblock Gallaudet University Press.

\bibitem[{McKee and Vale(2017)}]{mckee}
Rachel McKee and Mireille Vale. 2017.
\newblock \href {https://doi.org/10.1007/978-3-642-45369-4_34-1} {\emph{Sign
  Language Lexicography}}, pages 1--22.

\bibitem[{Meier et~al.(2002)Meier, Cormier, and
  Quinto-Pozos}]{texas2002modality}
R.P. Meier, K.~Cormier, and D.~Quinto-Pozos, editors. 2002.
\newblock \href {https://books.google.com/books?id=wkT8_WXozBsC}
  {\emph{Modality and Structure in Signed and Spoken Languages}}.
\newblock Cambridge University Press.

\bibitem[{M{\"u}ller et~al.(2023)M{\"u}ller, Alikhani, Avramidis
  et~al.}]{wmt_slt_23}
Mathias M{\"u}ller, Malihe Alikhani, Eleftherios Avramidis, et~al. 2023.
\newblock \href {https://doi.org/10.18653/v1/2023.wmt-1.4} {Findings of the
  second {WMT} shared task on sign language translation ({WMT}-{SLT}23)}.
\newblock In \emph{Proceedings of the Eighth Conference on Machine
  Translation}, pages 68--94, Singapore. Association for Computational
  Linguistics.

\bibitem[{M{\"u}ller et~al.(2022)M{\"u}ller, Ebling, Avramidis, Battisti,
  Berger, Bowden, Braffort, Cihan~Camg{\"o}z, Espa{\~n}a-bonet, Grundkiewicz,
  Jiang, Koller, Moryossef, Perrollaz, Reinhard, Rios, Shterionov,
  Sidler-miserez, and Tissi}]{wmt_slt}
Mathias M{\"u}ller, Sarah Ebling, Eleftherios Avramidis, Alessia Battisti,
  Mich{\`e}le Berger, Richard Bowden, Annelies Braffort, Necati
  Cihan~Camg{\"o}z, Cristina Espa{\~n}a-bonet, Roman Grundkiewicz, Zifan Jiang,
  Oscar Koller, Amit Moryossef, Regula Perrollaz, Sabine Reinhard, Annette
  Rios, Dimitar Shterionov, Sandra Sidler-miserez, and Katja Tissi. 2022.
\newblock \href {https://aclanthology.org/2022.wmt-1.71} {Findings of the first
  {WMT} shared task on sign language translation ({WMT}-{SLT}22)}.
\newblock In \emph{Proceedings of the Seventh Conference on Machine Translation
  (WMT)}, pages 744--772, Abu Dhabi, United Arab Emirates (Hybrid). Association
  for Computational Linguistics.

\bibitem[{Murray et~al.(2019)Murray, Hall, and Snoddon}]{signExposurePct}
Joseph~J Murray, Wyatte~C Hall, and Kristin Snoddon. 2019.
\newblock \href {https://doi.org/10.2471/BLT.19.229427} {Education and health
  of children with hearing loss: the necessity of signed languages.}

\bibitem[{Müller et~al.(2022)Müller, Jiang, Moryossef, Rios, and
  Ebling}]{muller2022considerations}
Mathias Müller, Zifan Jiang, Amit Moryossef, Annette Rios, and Sarah Ebling.
  2022.
\newblock \href {https://arxiv.org/abs/2211.15464} {Considerations for
  meaningful sign language machine translation based on glosses}.
\newblock \emph{Preprint}, arXiv:2211.15464.

\bibitem[{Müller et~al.(2019)Müller, Rios, Voita, and
  Sennrich}]{muller2019largescale}
Mathias Müller, Annette Rios, Elena Voita, and Rico Sennrich. 2019.
\newblock \href {https://arxiv.org/abs/1810.02268} {A large-scale test set for
  the evaluation of context-aware pronoun translation in neural machine
  translation}.
\newblock \emph{Preprint}, arXiv:1810.02268.

\bibitem[{Nagata and Morishita(2020)}]{nagata-morishita-2020-test}
Masaaki Nagata and Makoto Morishita. 2020.
\newblock \href {https://aclanthology.org/2020.lrec-1.457} {A test set for
  discourse translation from {J}apanese to {E}nglish}.
\newblock In \emph{Proceedings of the Twelfth Language Resources and Evaluation
  Conference}, pages 3704--3709, Marseille, France. European Language Resources
  Association.

\bibitem[{Nonaka et~al.(2015)Nonaka, Mesh, and Sagara}]{jslNameSigns}
Angela Nonaka, Kate Mesh, and Keiko Sagara. 2015.
\newblock \href {https://doi.org/10.1353/sls.2015.0025} {Signed names in
  japanese sign language: Linguistic and cultural analyses}.
\newblock \emph{Sign Language Studies}, 16(1):57–85.

\bibitem[{Padden(1986)}]{padden1986verbs}
Carol Padden. 1986.
\newblock Verbs and role-shifting in american sign language.
\newblock In \emph{Proceedings of the fourth national symposium on sign
  language research and teaching}, volume~44, page~57. National Association of
  the Deaf Silver Spring, MD.

\bibitem[{Papineni et~al.(2002)Papineni, Roukos, Ward, and
  Zhu}]{papineni-etal-2002-bleu}
Kishore Papineni, Salim Roukos, Todd Ward, and Wei-Jing Zhu. 2002.
\newblock \href {https://doi.org/10.3115/1073083.1073135} {{B}leu: a method for
  automatic evaluation of machine translation}.
\newblock In \emph{Proceedings of the 40th Annual Meeting of the Association
  for Computational Linguistics}, pages 311--318, Philadelphia, Pennsylvania,
  USA. Association for Computational Linguistics.

\bibitem[{Patrie and Johnson(2011)}]{patrie}
Carol~J Patrie and Robert~E Johnson. 2011.
\newblock \emph{{RSVP}: Fingerspelled word recognition through rapid serial
  visual presentation}.

\bibitem[{Post(2018)}]{post-2018-call}
Matt Post. 2018.
\newblock \href {https://www.aclweb.org/anthology/W18-6319} {A call for clarity
  in reporting {BLEU} scores}.
\newblock In \emph{Proceedings of the Third Conference on Machine Translation:
  Research Papers}, pages 186--191, Belgium, Brussels. Association for
  Computational Linguistics.

\bibitem[{Pu et~al.(2021)Pu, Chung, Parikh, Gehrmann, and
  Sellam}]{bleurt20paper}
Amy Pu, Hyung~Won Chung, Ankur~P Parikh, Sebastian Gehrmann, and Thibault
  Sellam. 2021.
\newblock Learning compact metrics for mt.
\newblock In \emph{Proceedings of EMNLP}.

\bibitem[{Raji and Buolamwini(2019)}]{auditing}
Inioluwa~Deborah Raji and Joy Buolamwini. 2019.
\newblock \href {https://doi.org/10.1145/3306618.3314244} {Actionable auditing:
  Investigating the impact of publicly naming biased performance results of
  commercial ai products}.
\newblock In \emph{Proceedings of the 2019 AAAI/ACM Conference on AI, Ethics,
  and Society}, AIES '19, page 429–435, New York, NY, USA. Association for
  Computing Machinery.

\bibitem[{Rendel et~al.(2018)Rendel, Bargones, Blake, Luetke, and
  Stryker}]{rendel2018signing}
Kabian Rendel, Jill Bargones, Britnee Blake, Barbara Luetke, and Deborah~S
  Stryker. 2018.
\newblock Signing exact english; a simultaneously spoken and signed
  communication option in deaf education.
\newblock \emph{Journal of Early Hearing Detection and Intervention},
  3(2):18--29.

\bibitem[{Sanabria et~al.(2018)Sanabria, Caglayan, Palaskar, Elliott, Barrault,
  Specia, and Metze}]{how2}
Ramon Sanabria, Ozan Caglayan, Shruti Palaskar, Desmond Elliott, Loïc
  Barrault, Lucia Specia, and Florian Metze. 2018.
\newblock \href {https://doi.org/10.48550/ARXIV.1811.00347} {How2: A
  large-scale dataset for multimodal language understanding}.
\newblock \emph{arXiv preprint}.

\bibitem[{Savoldi et~al.(2021)Savoldi, Gaido, Bentivogli, Negri, and
  Turchi}]{genderBias}
Beatrice Savoldi, Marco Gaido, Luisa Bentivogli, Matteo Negri, and Marco
  Turchi. 2021.
\newblock \href {https://doi.org/10.1162/tacl_a_00401} {{Gender Bias in Machine
  Translation}}.
\newblock \emph{Transactions of the Association for Computational Linguistics},
  9:845--874.

\bibitem[{Sellam et~al.(2020)Sellam, Das, and Parikh}]{sellam2020bleurt}
Thibault Sellam, Dipanjan Das, and Ankur~P Parikh. 2020.
\newblock Bleurt: Learning robust metrics for text generation.
\newblock In \emph{Proceedings of ACL}.

\bibitem[{Shen et~al.(2023)Shen, Yuan, Sheng, Du, and Yu}]{shen2023auslandaily}
Xin Shen, Shaozu Yuan, Hongwei Sheng, Heming Du, and Xin Yu. 2023.
\newblock \href {https://openreview.net/forum?id=g5v3Ig6WVq} {Auslan-daily:
  Australian sign language translation for daily communication and news}.
\newblock In \emph{Thirty-seventh Conference on Neural Information Processing
  Systems Datasets and Benchmarks Track}.

\bibitem[{Shi et~al.(2022)Shi, Brentari, Shakhnarovich, and Livescu}]{openasl}
Bowen Shi, Diane Brentari, Greg Shakhnarovich, and Karen Livescu. 2022.
\newblock \href {https://doi.org/10.48550/ARXIV.2205.12870} {Open-domain sign
  language translation learned from online video}.
\newblock \emph{arXiv preprint}.

\bibitem[{Shroyer and Shroyer(1984)}]{shroyer1984signs}
Edgar~H Shroyer and Susan~P Shroyer. 1984.
\newblock \emph{Signs across America: A look at regional differences in
  American Sign Language}.
\newblock Gallaudet University Press.

\bibitem[{Siddhant et~al.(2020)Siddhant, Bapna, Cao, Firat, Chen, Kudugunta,
  Arivazhagan, and Wu}]{siddhant-etal-2020-leveraging}
Aditya Siddhant, Ankur Bapna, Yuan Cao, Orhan Firat, Mia Chen, Sneha Kudugunta,
  Naveen Arivazhagan, and Yonghui Wu. 2020.
\newblock \href {https://doi.org/10.18653/v1/2020.acl-main.252} {Leveraging
  monolingual data with self-supervision for multilingual neural machine
  translation}.
\newblock In \emph{Proceedings of the 58th Annual Meeting of the Association
  for Computational Linguistics}, pages 2827--2835, Online. Association for
  Computational Linguistics.

\bibitem[{Sincan et~al.(2023)Sincan, Camgoz, and Bowden}]{sincan2023context}
Ozge~Mercanoglu Sincan, Necati~Cihan Camgoz, and Richard Bowden. 2023.
\newblock \href {https://arxiv.org/abs/2308.09622} {Is context all you need?
  scaling neural sign language translation to large domains of discourse}.
\newblock \emph{Preprint}, arXiv:2308.09622.

\bibitem[{Starner et~al.(2024)Starner, Forbes, So, Martin, Sridhar, Deshpande,
  Sepah, Shahryar, Bhardwaj, Kwok, Sehgal, Hassan, Neubauer, Vempala, Tan,
  Heath, Kumar, Mosur, Hall, Singh, Cui, Cameron, Dane, and Tanzer}]{popsign}
Thad Starner, Sean Forbes, Matthew So, David Martin, Rohit Sridhar, Gururaj
  Deshpande, Sam Sepah, Sahir Shahryar, Khushi Bhardwaj, Tyler Kwok, Daksh
  Sehgal, Saad Hassan, Bill Neubauer, Sofia~Anandi Vempala, Alec Tan, Jocelyn
  Heath, Unnathi~Utpal Kumar, Priyanka~Vijayaraghavan Mosur, Tavenner~M. Hall,
  Rajandeep Singh, Christopher~Zhang Cui, Glenn Cameron, Sohier Dane, and
  Garrett Tanzer. 2024.
\newblock Popsign asl v1.0: an isolated american sign language dataset
  collected via smartphones.
\newblock In \emph{Proceedings of the 37th International Conference on Neural
  Information Processing Systems}, NIPS '23, Red Hook, NY, USA. Curran
  Associates Inc.

\bibitem[{Stokoe~Jr(1969)}]{stokoe1969sign}
William~C Stokoe~Jr. 1969.
\newblock Sign language diglossia.

\bibitem[{Supalla and McKee(2002)}]{supalla2002role}
Sam Supalla and Cecile McKee. 2002.
\newblock The role of manually coded english in language development of deaf
  children.
\newblock \emph{Modality and structure in signed and spoken languages}, pages
  143--65.

\bibitem[{Supalla(1992)}]{namesigns}
Samuel~J. Supalla. 1992.
\newblock \emph{The Book of Name Signs: Naming in American Sign Language}.
\newblock DawnSignPress.

\bibitem[{Tarrés et~al.(2023)Tarrés, Gállego, Duarte, Torres, and
  i~Nieto}]{tarres2023sign}
Laia Tarrés, Gerard~I. Gállego, Amanda Duarte, Jordi Torres, and Xavier~Giró
  i~Nieto. 2023.
\newblock \href {https://arxiv.org/abs/2304.06371} {Sign language translation
  from instructional videos}.
\newblock \emph{Preprint}, arXiv:2304.06371.

\bibitem[{Thumann(2012)}]{thumann}
Mary Thumann. 2012.
\newblock \href {https://digitalcommons.unf.edu/joi/vol19/iss1/4/}
  {Fingerspelling in a word}.

\bibitem[{Uthus et~al.(2023)Uthus, Tanzer, and Georg}]{uthus2023youtubeasl}
David Uthus, Garrett Tanzer, and Manfred Georg. 2023.
\newblock \href {https://arxiv.org/abs/2306.15162} {Youtube-asl: A large-scale,
  open-domain american sign language-english parallel corpus}.
\newblock \emph{Preprint}, arXiv:2306.15162.

\bibitem[{Veale et~al.(1998)Veale, Conway, and Collins}]{veale1998challenges}
Tony Veale, Alan Conway, and Br{\'o}na Collins. 1998.
\newblock The challenges of cross-modal translation: English-to-sign-language
  translation in the zardoz system.
\newblock \emph{Machine Translation}, 13:81--106.

\bibitem[{Vicars(2023)}]{vicars2023alternating}
Bill Vicars. 2023.
\newblock \href {https://lifeprint.com/asl101/topics/alternating-diglossia.htm}
  {Alternating diglossia in the american deaf community: A dynamic interplay of
  {ASL} and english}.

\bibitem[{Voita et~al.(2019)Voita, Sennrich, and Titov}]{spokenContext}
Elena Voita, Rico Sennrich, and Ivan Titov. 2019.
\newblock \href {https://doi.org/10.18653/v1/P19-1116} {When a good translation
  is wrong in context: Context-aware machine translation improves on deixis,
  ellipsis, and lexical cohesion}.
\newblock In \emph{Proceedings of the 57th Annual Meeting of the Association
  for Computational Linguistics}, pages 1198--1212, Florence, Italy.
  Association for Computational Linguistics.

\bibitem[{Wager(2012)}]{wager}
Deborah~Stocks Wager. 2012.
\newblock \href {https://collections.lib.utah.edu/ark:/87278/s69p3gfz}
  {Fingerspelling in american sign language: A case study of styles and
  reduction}.

\bibitem[{Wang et~al.(2023)Wang, Lyu, Ji, Zhang, Yu, Shi, and
  Tu}]{wang-etal-2023-document-level}
Longyue Wang, Chenyang Lyu, Tianbo Ji, Zhirui Zhang, Dian Yu, Shuming Shi, and
  Zhaopeng Tu. 2023.
\newblock \href {https://doi.org/10.18653/v1/2023.emnlp-main.1036}
  {Document-level machine translation with large language models}.
\newblock In \emph{Proceedings of the 2023 Conference on Empirical Methods in
  Natural Language Processing}, pages 16646--16661, Singapore. Association for
  Computational Linguistics.

\bibitem[{Yin et~al.(2022)Yin, Zhao, Jin, Zhang, Zeng, and He}]{sp10}
Aoxiong Yin, Zhou Zhao, Weike Jin, Meng Zhang, Xingshan Zeng, and Xiaofei He.
  2022.
\newblock Mlslt: Towards multilingual sign language translation.
\newblock In \emph{Proceedings of the IEEE/CVF Conference on Computer Vision
  and Pattern Recognition (CVPR)}, pages 5109--5119.

\bibitem[{Yin et~al.(2021)Yin, Moryossef, Hochgesang, Goldberg, and
  Alikhani}]{kayo}
Kayo Yin, Amit Moryossef, Julie Hochgesang, Yoav Goldberg, and Malihe Alikhani.
  2021.
\newblock \href {https://doi.org/10.18653/v1/2021.acl-long.570} {Including
  signed languages in natural language processing}.
\newblock In \emph{Proceedings of the 59th Annual Meeting of the Association
  for Computational Linguistics and the 11th International Joint Conference on
  Natural Language Processing (Volume 1: Long Papers)}, pages 7347--7360,
  Online. Association for Computational Linguistics.

\bibitem[{Zhao et~al.(2000)Zhao, Kipper, Schuler, Vogler, Badler, and
  Palmer}]{earlySlt}
Liwei Zhao, Karin Kipper, William Schuler, Christian Vogler, Norman Badler, and
  Martha Palmer. 2000.
\newblock A machine translation system from english to american sign language.
\newblock In \emph{Envisioning Machine Translation in the Information Future},
  pages 54--67, Berlin, Heidelberg. Springer Berlin Heidelberg.

\end{thebibliography}
\bibliographystyle{acl_natbib}

\appendix

\section{Human Annotations in Sign Language Translation Datasets}
\label{sec:related-work-expanded}

In this section we provide more analysis of the human annotations used to construct a variety of sign language translation datasets:

\begin{itemize}
    \item Content4All~\cite{content4all} is a collection of news broadcasts interpreted into Swiss German Sign Language (DSGS) and Flemish Sign Language. The broadcasts contain weakly aligned captions by construction, and human annotators manually align a subset of captions with discourse-level context.
    \item The WMT-SLT datasets~\cite{wmt_slt,wmt_slt_23} are built on several sources of news broadcasts in Swiss German Sign Language, some produced in DSGS and others interpreted. Competition entries are rated by humans, and the reference translations are scored in the same human evaluation framework as a baseline, but ``human translation'' and ``reference translation'' are treated interchangeably. WMT-SLT23 finds that the references in one test set are rated worse than the others, and raises the possibility that this is related to discourse context but does not explore it further.
    \item BOBSL~\citep{bobsl} is a dataset composed of BBC programs interpreted into British Sign Language. Human annotators are used to evaluate preprocessing decisions and clean up the test set.
    \item How2Sign~\cite{how2sign} is an American Sign Language dataset containing studio translations of ``how to'' videos. Human annotations are used to align captions and evaluate the intelligibility of skeletons vs. generated videos.
    \item OpenASL~\cite{openasl} is an American Sign Language dataset consisting of videos mined from several YouTube channels. Human ratings are only used to evaluate how well the caption tracks attached to these videos are aligned to their content.
    \item ISLTranslate~\cite{joshi2023isltranslate} is built from children's educational content produced in Indian Sign Language. A signer performs a human baseline given full discourse context to validate the quality of the reference captions, not to sanity check the task framing.
    \item Auslan-Daily~\citep{shen2023auslandaily} is a dataset composed of of Australian Sign Language TV programs. Human experts are used to perform fine-grained annotations and check each other's work given full video context, but not check the task framing itself.
    \item YouTube-ASL~\citep{uthus2023youtubeasl} is a corpus of captioned American Sign Language videos drawn from YouTube. Human annotators are used only to filter out videos with low-quality signing or captions.
    \item JWSign~\citep{gueuwou2023jwsign} is a dataset of Bible translations into many sign languages. No human annotators were used when constructing the dataset, since it is constructed from preexisting clean data.
\end{itemize}

The fingerspelling recognition (not sign language translation) datasets ChicagoFSWild~\citep{fs18slt} and ChicagoFSWild+~\citep{fs18iccv}, which consist of clips extracted from continuous signing data, do provide references for human performance within the clip-level task framing. They observe that the baseline scores are lower than inter-annotator agreement between the ground truth annotators (who had access to the surrounding video), meaning that something is lost without context. This task has even less context than sentence-level translation, and could be seen as a manifestation of rapid fingerspelling, described in Section~\ref{sec:oov-terms}. However, it is not clear whether the ground truth annotators had access to captions, which could improve results beyond what is actually possible given the entire video (but only the video) as context (like the \textbf{s$_{i-1:i}$}, \textbf{t$_{i-1}$} and \textbf{s$_{i-1: i}$}, \textbf{t$_{0:i-1}$} settings in our How2Sign human baseline).

\section{\emph{How2Sign} Human Baseline}

\subsection{Annotator Instructions}
\label{app:human-baseline-instructions}

""" \\
For each video id (sentence) there are 4 experimental conditions:

\begin{enumerate}
\item Translate from a source clip
\item Translate from a source clip, extended backwards in time to include the previous sentence as context
\item Translate from the above clip, but also with the ground truth English translation for the previous sentence as context
\item Translate from the above clip, but also with the ground truth English translation for the entire narrative up to that point as context
\end{enumerate}

Each of those gives strictly more context than the previous one, so it should be legitimate for a single person to do all of them in sequence for a single sentence. But that means it's important that you don't see the extra context too soon. This is why certain cells are redacted (filled in with black). You can unredact the cell by resetting the fill.

So for each sentence/video id, you should do the following:

\begin{enumerate}
\item Open the first video link. This is a clip containing only the sentence in question. Translate it into English and write the result in the first row under "your translation goes here".
\item Open the second video link. This clip also includes the sentence before the sentence in question. Use this extra context to improve your translation of the sentence in question (if it makes a difference) and write it in the second row under "your translation goes here", but do not translate the extra sentence included in the video. It's just for context.
\item Using the same video link (second), reveal the contents of the first context cell. This is the English translation of the previous sentence (the one included in the extended video). Use this extra context to improve your translation (if it makes a difference) and write it in the third row.
\item Using the same video link, reveal the contents of the second context cell. This is the English translation of the entire narrative up to this point. Use this extra context to improve your translation (if it makes a difference) and write it in the fourth row. (In some cases, the narrative up to this point only consists of the previous sentence, so \#3 and \#4 have exactly the same context. Just copy/paste your translation from above for this case.)
\end{enumerate}

Afterwards, you can reveal the ground truth sentence. There are three more annotations that I'd like to get (put it on the same row as the ground truth sentence):
\begin{enumerate}
\item How well could you understand the sentence in isolation? Pick one of {"not at all", "somewhat", "mostly", "completely"}
\item Is the clip signed in natural ASL? Pick one of {"no", "eh", "yes"}. (For example, SEE would be considered "no". PSE might be considered "meh".)
\item Is this an interesting example? You can leave a note here if this sentence might be an interesting example for the paper (i.e. it depends on long term context in a way that is interesting/exemplary)
\end{enumerate}
As a general note: when you translate, if there is ambiguity just give your best guess. Pretend that you're confident (though you might hedge by using pronouns, etc.). This is necessary in order to get a like-for-like comparison with machine translation results.

Let me know if you have any questions (or if any of the clips seem misaligned, links are broken, etc.). \\

PS: Here is a sample of sentences from the dataset so you can get a sense of the style/tone for your translations. It's drawn from a collection of "how to" instructional narratives.
\begin{itemize}
\item My name is Daniel King, and I'm an experienced pattern maker, designer and sewer.
\item So thanks a lot for joining us here, I appreciate it.
\item There's an old saying that I think is real important to remember when we're talking about criticism, whether it's written or whether it's spoken.
\item But the most important thing is by using your legs, a lot of time you see players come up and shoot their free throw and they stay flat footed and then end up hitting the ball on the front of the rim.
\item Sometimes it gets a little stuck, always wipe the edge though of your exacto blade off, that blade is going to end up tending to be a blade that your not really going to be able to use for cutting much anymore, so you may want to have two of the tools available to you so that in case one of them, you want to just keep that open for cutting and the other one you can use for lifting the materials up when they get stuck.
\item Fold this bottom up to the center, like so.
\item I want to form an after school program that involves at risk teens be able to overcome their differences so that we can bridge the gaps of our society and our future.
\end{itemize}
"""

\subsection{Baseline Results}
\label{app:human-baseline-results}

See Table~\ref{tab:baseline} for the complete set of translations comprising our human baseline.

\begin{table*}[]
    \caption{\textbf{Complete set of translations comprising our human baseline}, alphabetized by video id. ``-'' means that the translation is the same as in the previous setting.}
    \centering
    \small
    \begin{tabular}{cccp{9.2cm}}
        \toprule
        \textbf{video id} & \textbf{interpreter} & \textbf{setting} & \textbf{translation} \\
        \midrule
-fZc293MpJk-1 & A & ground truth & By moving the stick, you cause pressure to increase or decrease the angle of attack on that particular raising or lowering the wing. \\ & & \textbf{s$_i$} & That causes pressure, when moving the joystick side to side, it rocks side to side on the surface. \\ & & \textbf{s$_{i-1:i}$} & That causes pressure, moving the joystick side to side makes the wings rock side to side. \\ & & \textbf{s$_{i-1:i}$}, \textbf{t$_{i-1}$} & - \\ & & \textbf{s$_{i-1: i}$}, \textbf{t$_{0:i-1}$} & - \\ \midrule
-g0iPSnQt6w-1 & A & ground truth & And I'm actually going to lock my wrists when I pike. \\ & & \textbf{s$_i$} & Introduce wrist clamp, locked clamp. \\ & & \textbf{s$_{i-1:i}$} & Underneath, leg clamped, locked clamp. \\ & & \textbf{s$_{i-1:i}$}, \textbf{t$_{i-1}$} & - \\ & & \textbf{s$_{i-1: i}$}, \textbf{t$_{0:i-1}$} & - \\ \midrule
-g0sqksgyc4-2 & B & ground truth & In boxing you always want to be trying to be moving forward, you want to be trying to be pushed to fight, always trying to be moving forward. \\ & & \textbf{s$_i$} & Boxers always want to try to move closer, you want to try to push the fight, try to move closer. \\ & & \textbf{s$_{i-1:i}$} & - \\ & & \textbf{s$_{i-1:i}$}, \textbf{t$_{i-1}$} & - \\ & & \textbf{s$_{i-1: i}$}, \textbf{t$_{0:i-1}$} & - \\ \midrule
-g45vqccdzI-1 & A & ground truth & And we can get a little bit of a jump here and here we are on the other side of that door. \\ & & \textbf{s$_i$} & Riding on it, when you arrive, jump into it. \\ & & \textbf{s$_{i-1:i}$} & - \\ & & \textbf{s$_{i-1:i}$}, \textbf{t$_{i-1}$} & We ride on it, and when we arrive, we jump into the portal. \\ & & \textbf{s$_{i-1: i}$}, \textbf{t$_{0:i-1}$} & We ride on the transport plate t oreach the place where we can jump in. \\ \midrule
37ZtKNf6Yd8-1 & A & ground truth & Now the tuning of this instrument, you have the same string on the top and bottom and then you have a three and a five of the mayor scale on the inside of the instrument. \\ & & \textbf{s$_i$} & Hear drums, hear guitar, top and bottom same. You have three to five ayer between scalex inside things. \\ & & \textbf{s$_{i-1:i}$} & - \\ & & \textbf{s$_{i-1:i}$}, \textbf{t$_{i-1}$} & - \\ & & \textbf{s$_{i-1: i}$}, \textbf{t$_{0:i-1}$} & Hear adjustments, listen to a few strums, adjustments at top or bottom have same effect, three to five major between scales inside things. \\ \midrule
3ddzkmFPEBU-1 & A & ground truth & One would be a string winder, which is used on the tuning machine to wind it as you're putting the string on, make it much quicker than turning by hand. \\ & & \textbf{s$_i$} & A string winder helps tune machine guitar, it will help adjust tune while you listen - will help do it faster than winding at the end, meh. \\ & & \textbf{s$_{i-1:i}$} & One is a string winder that helps machine-tune a guitar, it will help adjust tune while you listen - will help do it faster than hand-winding at the guitar's end, no need for that. \\ & & \textbf{s$_{i-1:i}$}, \textbf{t$_{i-1}$} & - \\ & & \textbf{s$_{i-1: i}$}, \textbf{t$_{0:i-1}$} & - \\ \midrule
8kAWy2YodzQ-1 & A & ground truth & And checking out the second one. \\ & & \textbf{s$_i$} & Playing guitar. \\ & & \textbf{s$_{i-1:i}$} & Testing a couple of strokes on guitar. \\ & & \textbf{s$_{i-1:i}$}, \textbf{t$_{i-1}$} & - \\ & & \textbf{s$_{i-1: i}$}, \textbf{t$_{0:i-1}$} & Test a couple of strokes on one string of the ukulele. \\ \midrule
92V3oH63zbQ-1 & A & ground truth & We've talked about hitting inside pitches. \\ & & \textbf{s$_i$} & Now inside do clomp. \\ & & \textbf{s$_{i-1:i}$} & Now inside cast the fishing line. \\ & & \textbf{s$_{i-1:i}$}, \textbf{t$_{i-1}$} & Now inside is one kind of baseball pitch. \\ & & \textbf{s$_{i-1: i}$}, \textbf{t$_{0:i-1}$} & - \\ \midrule
FZCF7kPIyOk-1 & A & ground truth & It's not really going to add to the reception of your script. \\ & & \textbf{s$_i$} & I don't know... maybe that will help people listen and accept something on credit. \\ & & \textbf{s$_{i-1:i}$} & - \\ & & \textbf{s$_{i-1:i}$}, \textbf{t$_{i-1}$} & - \\ & & \textbf{s$_{i-1: i}$}, \textbf{t$_{0:i-1}$} & I'm not sure, but this advice should help people listen and accept your script. \\

        \bottomrule
    \end{tabular}

    \label{tab:baseline}
\end{table*}

\begin{table*}[]
    \centering
    \small
    \begin{tabular}{cccp{9.2cm}}
        \toprule
        \textbf{video id} & \textbf{interpreter} & \textbf{setting} & \textbf{translation} \\
        \midrule
FZLxEwsoc1c-8 & C & ground truth & That's basically the explanation of a nap. \\ & & \textbf{s$_i$} & That's the basic explanation of an app. \\ & & \textbf{s$_{i-1:i}$} & That's the basic explanation of an ap. \\ & & \textbf{s$_{i-1:i}$}, \textbf{t$_{i-1}$} & That's the basic explanation of a nap. \\ & & \textbf{s$_{i-1: i}$}, \textbf{t$_{0:i-1}$} & - \\ \midrule
FZNuNG9UBnw-1 & A & ground truth & Hi, I'm Captain Joe Bruni, and what I want to talk about is how to visually identify prescription drugs. \\ & & \textbf{s$_i$} & I'm Captain Ernie, I want to discuss how to visually identify an Rx drug. \\ & & \textbf{s$_{i-1:i}$} & - \\ & & \textbf{s$_{i-1:i}$}, \textbf{t$_{i-1}$} & - \\ & & \textbf{s$_{i-1: i}$}, \textbf{t$_{0:i-1}$} & - \\ \midrule
FZbyRzy4huk-8 & C & ground truth & We cook it for 1 1/2 to 2 hours once it's ready you can see it's nice and soft, we're going to drain it and we're going to continue to the next step. \\ & & \textbf{s$_i$} & Give it an hour and a half to two hours, when it looks ready, it will be soft and heavy, then drain the water, then continue to the next one. \\ & & \textbf{s$_{i-1:i}$} & - \\ & & \textbf{s$_{i-1:i}$}, \textbf{t$_{i-1}$} & - \\ & & \textbf{s$_{i-1: i}$}, \textbf{t$_{0:i-1}$} & - \\ \midrule
FZd8Iv9ACVw-8 & C & ground truth & Ok, first of all we can demonstrate with two of these snow tires the little different, the little cuts in the tires are called sipes. \\ & & \textbf{s$_i$} & Two different ones have red cuts called spies. \\ & & \textbf{s$_{i-1:i}$} & - \\ & & \textbf{s$_{i-1:i}$}, \textbf{t$_{i-1}$} & Two different snow tires, one has red cuts called spikes. \\ & & \textbf{s$_{i-1: i}$}, \textbf{t$_{0:i-1}$} & - \\ \midrule
FZrU\_mEryAs-2 & B & ground truth & That's a really good way for a child, a younger child to be able to point to a stranger how they can contact you. \\ & & \textbf{s$_i$} & Good way for a younger child to show a stranger their bracelet so they can contact you. \\ & & \textbf{s$_{i-1:i}$} & - \\ & & \textbf{s$_{i-1:i}$}, \textbf{t$_{i-1}$} & - \\ & & \textbf{s$_{i-1: i}$}, \textbf{t$_{0:i-1}$} & - \\ \midrule
FZrWOf-oGDk-8 & C & ground truth & And then either continue back to the back of the hook or up to the front. \\ & & \textbf{s$_i$} &  \\ & & \textbf{s$_{i-1:i}$} & - \\ & & \textbf{s$_{i-1:i}$}, \textbf{t$_{i-1}$} & - \\ & & \textbf{s$_{i-1: i}$}, \textbf{t$_{0:i-1}$} & - \\ \midrule
Fz-N1S0swh8-8 & C & ground truth & I loved it, it actually tasted really good. \\ & & \textbf{s$_i$} & Spanish ox \\ & & \textbf{s$_{i-1:i}$} & - \\ & & \textbf{s$_{i-1:i}$}, \textbf{t$_{i-1}$} & Spanish or Mexican \\ & & \textbf{s$_{i-1: i}$}, \textbf{t$_{0:i-1}$} & - \\ \midrule
FzAIlhumvMA-2 & B & ground truth & There's also information on here about mailing the sample to a laboratory for confidential confirmation. \\ & & \textbf{s$_i$} & Also the information here - mailed sample to the lab for confidentiality - promise. \\ & & \textbf{s$_{i-1:i}$} & - \\ & & \textbf{s$_{i-1:i}$}, \textbf{t$_{i-1}$} & - \\ & & \textbf{s$_{i-1: i}$}, \textbf{t$_{0:i-1}$} & - \\ \midrule
FzOQMA-CVPc-2 & B & ground truth & So, just try and come up with a budget for your party and you want to have this much money for food and for decorations and just split it up. \\ & & \textbf{s$_i$} & Just try to come first budget for party you want to have this much money for food and for decorations, split. \\ & & \textbf{s$_{i-1:i}$} & - \\ & & \textbf{s$_{i-1:i}$}, \textbf{t$_{i-1}$} & Just try to first come up with a budget for the party - you want to have this much money for food and for decorations, split. \\ & & \textbf{s$_{i-1: i}$}, \textbf{t$_{0:i-1}$} & - \\ \midrule
FzQPg4aqNYc-1 & A & ground truth & That's how I serve that. \\ & & \textbf{s$_i$} & How I give tea to the customer. \\ & & \textbf{s$_{i-1:i}$} & That's how I give tea to the customer. \\ & & \textbf{s$_{i-1:i}$}, \textbf{t$_{i-1}$} & - \\ & & \textbf{s$_{i-1: i}$}, \textbf{t$_{0:i-1}$} & That's how I stick in the leaf and then give it to the customer. \\
        \bottomrule
    \end{tabular}
    \label{tab:baseline2}
\end{table*}

\begin{table*}[]
    \centering
    \small
    \begin{tabular}{cccp{9.2cm}}
        \toprule
        \textbf{video id} & \textbf{interpreter} & \textbf{setting} & \textbf{translation} \\
        \midrule
FzUdcaxw\_vs-2 & B & ground truth & Come on; let's get ironing. \\ & & \textbf{s$_i$} & Come on, just iron it. \\ & & \textbf{s$_{i-1:i}$} & - \\ & & \textbf{s$_{i-1:i}$}, \textbf{t$_{i-1}$} & - \\ & & \textbf{s$_{i-1: i}$}, \textbf{t$_{0:i-1}$} & - \\ \midrule
FzWvE\_\_PamM-2 & B & ground truth & Another great way to spruce your page is add video to your page and to do that you want to look to the top right. \\ & & \textbf{s$_i$} & The other way to show adding a video to a page. Move your video over to your page, then we want to look at the top right. \\ & & \textbf{s$_{i-1:i}$} & - \\ & & \textbf{s$_{i-1:i}$}, \textbf{t$_{i-1}$} & - \\ & & \textbf{s$_{i-1: i}$}, \textbf{t$_{0:i-1}$} & - \\ \midrule
FzaQ-Q5gSmI-1 & A & ground truth & And this is the base plate. \\ & & \textbf{s$_i$} & On the bottom it has a strip called the base. It's a plate. \\ & & \textbf{s$_{i-1:i}$} & - \\ & & \textbf{s$_{i-1:i}$}, \textbf{t$_{i-1}$} & The bottom of the saw has a strip called the base. It's a plate. \\ & & \textbf{s$_{i-1: i}$}, \textbf{t$_{0:i-1}$} & - \\ \midrule
Fzj3jz2Imf0-1 & A & ground truth & We're going to do some hand vibrato exercises, finger vibrato exercises, as well as arm vibrato exercises so that you can decide for yourself which type of vibrato you'd like to use and you'll have all the information that you need to get started. \\ & & \textbf{s$_i$} & Notice we'll discuss this more, like practice hand vibrato - also practice arm vibrato -- then decide for yourself which you prefer, it's important to have all the information needed to start playing violin. \\ & & \textbf{s$_{i-1:i}$} & - \\ & & \textbf{s$_{i-1:i}$}, \textbf{t$_{i-1}$} & - \\ & & \textbf{s$_{i-1: i}$}, \textbf{t$_{0:i-1}$} & - \\ \midrule
FzmL8SL6Bow-8 & C & ground truth & So just go in and you can even take this guy here and go in there, flatten it down, real nice. \\ & & \textbf{s$_i$} & Take a scoop of it, put it in it, that helps shape it to be a square box and flat. \\ & & \textbf{s$_{i-1:i}$} & Take a scoop of clay, put it in the bowl, that helps flesh out the shape and make the bottom flat. \\ & & \textbf{s$_{i-1:i}$}, \textbf{t$_{i-1}$} & - \\ & & \textbf{s$_{i-1: i}$}, \textbf{t$_{0:i-1}$} & - \\ \midrule
FzoUVr98JmQ-8 & C & ground truth & Salt, about 1 teaspoon full, add a little bit of chili powder; it depends if you want it very spicy, you can go for more. \\ & & \textbf{s$_i$} & Around a teaspoon of salt, add some chili flakes, if you like it hot, add more. \\ & & \textbf{s$_{i-1:i}$} & - \\ & & \textbf{s$_{i-1:i}$}, \textbf{t$_{i-1}$} & - \\ & & \textbf{s$_{i-1: i}$}, \textbf{t$_{0:i-1}$} & - \\ \midrule
G-0gYel1YA8-2 & B & ground truth & Some people get kind of confused about the time that it takes for their piercings to close up because most people are used to having their eyes pierced and they're used to having them pierced for a long time and those most of the time don't close up. \\ & & \textbf{s$_i$} & Why do piercings close up? People are confused about the timing when you take piercings off, because people tend to have... \\ & & \textbf{s$_{i-1:i}$} & Why piercings close up. Some people get confused about the timing when removing their piercings because people tend to have... \\ & & \textbf{s$_{i-1:i}$}, \textbf{t$_{i-1}$} & - \\ & & \textbf{s$_{i-1: i}$}, \textbf{t$_{0:i-1}$} & - \\ \midrule
G05uFub3YFc-2 & B & ground truth & We're going to drop that elbow down as we lift the top arm up at least to the ceiling, and if you're feeling really open and really comfortable with this pose you can reach it up alongside the ear, but don't let the shoulders creep up. \\ & & \textbf{s$_i$} & Keep one elbow down and bring the other one around above your head, at least try to touch the ceiling. If you feel really comfortable, you can stretch futher, but keep your neck loose, don't squeeze your arm to your ear. \\ & & \textbf{s$_{i-1:i}$} & Keeping that elbow down, move the other one around above your head, at least try to touch the ceiling. If you feel really comfortable, you can stretch futher, but keep your neck loose, don't squeeze your arm to your ear. \\ & & \textbf{s$_{i-1:i}$}, \textbf{t$_{i-1}$} & - \\ & & \textbf{s$_{i-1: i}$}, \textbf{t$_{0:i-1}$} & - \\
        \bottomrule
    \end{tabular}
    \label{tab:baseline3}
\end{table*}

\begin{table*}[]
    \centering
    \small
    \begin{tabular}{cccp{9.2cm}}
        \toprule
        \textbf{video id} & \textbf{interpreter} & \textbf{setting} & \textbf{translation} \\
        \midrule
G06Irzcwxiw-1 & A & ground truth & You enjoy the moment of what you're doing. \\ & & \textbf{s$_i$} & Need to enjoy that moment. \\ & & \textbf{s$_{i-1:i}$} & You need to enjoy each moment. \\ & & \textbf{s$_{i-1:i}$}, \textbf{t$_{i-1}$} & - \\ & & \textbf{s$_{i-1: i}$}, \textbf{t$_{0:i-1}$} & - \\ \midrule
G095RWKQ39g-1 & A & ground truth & But as you can see, because of the small size, if I'm going to use this red dot finder and I'm under six foot, I've got to get down here and locate my objects. \\ & & \textbf{s$_i$} & There's a little red divider that says 6ft. Do I still have to bend under it to see it? I don't know. \\ & & \textbf{s$_{i-1:i}$} & There's a little red dot finder. I'm 6 feet tall, so do I still have to bend down to see the crosshairs? I'm not sure. \\ & & \textbf{s$_{i-1:i}$}, \textbf{t$_{i-1}$} & - \\ & & \textbf{s$_{i-1: i}$}, \textbf{t$_{0:i-1}$} & - \\ \midrule
G0MjvzT\_UqM-2 & B & ground truth & Ready, inhale. \\ & & \textbf{s$_i$} & Ready? Breathe in your kee. \\ & & \textbf{s$_{i-1:i}$} & Ready? Breathe in. Your knee - breathe in. \\ & & \textbf{s$_{i-1:i}$}, \textbf{t$_{i-1}$} & - \\ & & \textbf{s$_{i-1: i}$}, \textbf{t$_{0:i-1}$} & - \\ \midrule
G0PNAsonBGk-2 & B & ground truth & Now we're going to turn, instead of bringing the hand up, we leave the hand down, just like this. \\ & & \textbf{s$_i$} & Now turn your hands up - like, leave your hands down, like this. \\ & & \textbf{s$_{i-1:i}$} & - \\ & & \textbf{s$_{i-1:i}$}, \textbf{t$_{i-1}$} & - \\ & & \textbf{s$_{i-1: i}$}, \textbf{t$_{0:i-1}$} & - \\ \midrule
G0Q6AlvH96I-2 & B & ground truth & Here, two, three, four, elbow and follow wherever you're going to go, like the knee to the groin and your elbow. \\ & & \textbf{s$_i$} & Here, two, three, four, elbow follows you wherever you go, like your knee or organs, your elbow. \\ & & \textbf{s$_{i-1:i}$} & - \\ & & \textbf{s$_{i-1:i}$}, \textbf{t$_{i-1}$} & - \\ & & \textbf{s$_{i-1: i}$}, \textbf{t$_{0:i-1}$} & - \\ \midrule
G19uBylwQww-2 & B & ground truth & Hi, my name is Robert Segundo and today I'm going to teach you how to make one of my favorite paper airplanes, the simple one. \\ & & \textbf{s$_i$} & My name is Robert Segundo and today is about expert community, my favorite way to play is making paper airplanes. \\ & & \textbf{s$_{i-1:i}$} & My name is Robert Segundo and today is about the expert community, my favorite way to play is making paper airplanes. \\ & & \textbf{s$_{i-1:i}$}, \textbf{t$_{i-1}$} & - \\ & & \textbf{s$_{i-1: i}$}, \textbf{t$_{0:i-1}$} & - \\ \midrule
G1GUMky8kWc-2 & B & ground truth & One and two and three and four. \\ & & \textbf{s$_i$} & And one, two, three, four. And one. \\ & & \textbf{s$_{i-1:i}$} & - \\ & & \textbf{s$_{i-1:i}$}, \textbf{t$_{i-1}$} & - \\ & & \textbf{s$_{i-1: i}$}, \textbf{t$_{0:i-1}$} & - \\ \midrule
G1LiGqM3FhM-8 & C & ground truth & If you wanted to do something minor, you could make cross cuts like this. \\ & & \textbf{s$_i$} & If you want something small, you can make roosevelt to show. \\ & & \textbf{s$_{i-1:i}$} & If you want something small, you can make roosevelt for example. \\ & & \textbf{s$_{i-1:i}$}, \textbf{t$_{i-1}$} & If you want something small, you can make a crosscut for example. \\ & & \textbf{s$_{i-1: i}$}, \textbf{t$_{0:i-1}$} & - \\ \midrule
G1QiXuldOxM-8 & C & ground truth & Make sure you're wedging properly. \\ & & \textbf{s$_i$} & Look. \\ & & \textbf{s$_{i-1:i}$} & - \\ & & \textbf{s$_{i-1:i}$}, \textbf{t$_{i-1}$} & - \\ & & \textbf{s$_{i-1: i}$}, \textbf{t$_{0:i-1}$} & - \\

        \bottomrule
    \end{tabular}
    \label{tab:baseline4}
\end{table*}

\begin{table*}[]
    \centering
    \small
    \begin{tabular}{cccp{9.2cm}}
        \toprule
        \textbf{video id} & \textbf{interpreter} & \textbf{setting} & \textbf{translation} \\
        \midrule
G1hb5HugzVk-8 & C & ground truth & A nice item to serve with that spaghetti would be a green salad and maybe some garlic bread, a nice simple garlic receipt would be to take some butter and mix it with some garlic salt or garlic powder but you want that salty that is in there. \\ & & \textbf{s$_i$} & Nice things, maybe put the pasta and the green salad, maybe garlic bread, it's a simple garlic recipe: butter, garlic salt or garlic powder. You want that salty taste. \\ & & \textbf{s$_{i-1:i}$} & We can have some nice spaghetti and green salad. Maybe garlic bread, it's a simple recipe: butter and garlic salt or garlic powder - you want it to have that salty taste. \\ & & \textbf{s$_{i-1:i}$}, \textbf{t$_{i-1}$} & - \\ & & \textbf{s$_{i-1: i}$}, \textbf{t$_{0:i-1}$} & - \\ \midrule
G1jsDl1mVvk-1 & A & ground truth & Sometimes it could be, you know, the black and white stripes. \\ & & \textbf{s$_i$} & Sometimes you can have a solid color shirt with stripes. \\ & & \textbf{s$_{i-1:i}$} & Sometimes they will have a solid color shirt with stripes on it. \\ & & \textbf{s$_{i-1:i}$}, \textbf{t$_{i-1}$} & - \\ & & \textbf{s$_{i-1: i}$}, \textbf{t$_{0:i-1}$} & - \\ \midrule
G1lNlhjWC1I-8 & C & ground truth & But one other tip when choosing eye shadow color is actually take a look at color of there eyes. \\ & & \textbf{s$_i$} & A tip when picking the color of your eyeshadow - really look at the color of your eyes. \\ & & \textbf{s$_{i-1:i}$} & - \\ & & \textbf{s$_{i-1:i}$}, \textbf{t$_{i-1}$} & - \\ & & \textbf{s$_{i-1: i}$}, \textbf{t$_{0:i-1}$} & - \\ \midrule
G1nq4fYZiyQ-8 & C & ground truth & She's going to take this it reach forward press firmly into the outer edges of the block and with her inhale, she's going to reach her arms up. \\ & & \textbf{s$_i$} & Go ahead and press it firmly on either side. At the same time breathe in, and it will grow in height. \\ & & \textbf{s$_{i-1:i}$} & - \\ & & \textbf{s$_{i-1:i}$}, \textbf{t$_{i-1}$} & - \\ & & \textbf{s$_{i-1: i}$}, \textbf{t$_{0:i-1}$} & Go ahead and press it firmly on either side. At the same time breathe in, and bring your arms above your head. \\ \midrule
G21Gx\_C18IA-2 & B & ground truth & Once again, this is Gabriela Garzon at G.G. \\ & & \textbf{s$_i$} & Once again my name is Gabriel La Garrlon, or G.G. \\ & & \textbf{s$_{i-1:i}$} & - \\ & & \textbf{s$_{i-1:i}$}, \textbf{t$_{i-1}$} & - \\ & & \textbf{s$_{i-1: i}$}, \textbf{t$_{0:i-1}$} & Once again, my name is Gabriela Garzon with G.G. \\ \midrule
G23JltC2N8g-5 & D & ground truth & But for safety purposes if that's necessary bring yourself against the wall, and bring yourself right back, and bring your feet up. \\ & & \textbf{s$_i$} & Core of your body - the center part of your body, but you're not using it well, but it's for safety. \\ & & \textbf{s$_{i-1:i}$} & - \\ & & \textbf{s$_{i-1:i}$}, \textbf{t$_{i-1}$} & - \\ & & \textbf{s$_{i-1: i}$}, \textbf{t$_{0:i-1}$} & - \\ \midrule
G2Go6a76xd0-5 & D & ground truth & You need to consider whether the horse has an illness or an injury. \\ & & \textbf{s$_i$} & Consider if either of your horses have illness or injury. \\ & & \textbf{s$_{i-1:i}$} & Consider whether your horses have illness or injury. \\ & & \textbf{s$_{i-1:i}$}, \textbf{t$_{i-1}$} & - \\ & & \textbf{s$_{i-1: i}$}, \textbf{t$_{0:i-1}$} & - \\ \midrule
G2VAlFdgof4-5 & D & ground truth & That is how we do the second line in our heart pulse and monitor design. \\ & & \textbf{s$_i$} & How are we doing the second line in our heart pulse and monitor design. \\ & & \textbf{s$_{i-1:i}$} & - \\ & & \textbf{s$_{i-1:i}$}, \textbf{t$_{i-1}$} & - \\ & & \textbf{s$_{i-1: i}$}, \textbf{t$_{0:i-1}$} & - \\ \midrule
G2dND014Ps4-5 & D & ground truth & This lever is very important when you want to open up your scooter because you can't ride it like this. \\ & & \textbf{s$_i$} & Really important - you want to open up your scooter because you can't ride like this. \\ & & \textbf{s$_{i-1:i}$} & That's really important. So I want you to open up your scooter because you can't ride it like this. \\ & & \textbf{s$_{i-1:i}$}, \textbf{t$_{i-1}$} & - \\ & & \textbf{s$_{i-1: i}$}, \textbf{t$_{0:i-1}$} & - \\

        \bottomrule
    \end{tabular}
    \label{tab:baseline5}
\end{table*}

\begin{table*}[]
    \centering
    \small
    \begin{tabular}{cccp{9.2cm}}
        \toprule
        \textbf{video id} & \textbf{interpreter} & \textbf{setting} & \textbf{translation} \\
        \midrule

G2hnUeetWcc-5 & D & ground truth & Look up. \\ & & \textbf{s$_i$} & Look up. \\ & & \textbf{s$_{i-1:i}$} & - \\ & & \textbf{s$_{i-1:i}$}, \textbf{t$_{i-1}$} & - \\ & & \textbf{s$_{i-1: i}$}, \textbf{t$_{0:i-1}$} & - \\ \midrule
G2lEchCCRAo-5 & D & ground truth & So you can see in comparison in size they are comfortable so when you are looking at a teapot or a sugar bowl with a set like this you want to make sure that the sizes are appropriate for what you are buying. \\ & & \textbf{s$_i$} & See that they are comparable in size and that they are comfortable, so when you are looking at teapots or sugar bowl sets like this you want to make sure that the sizes are right for what you are buying. \\ & & \textbf{s$_{i-1:i}$} & - \\ & & \textbf{s$_{i-1:i}$}, \textbf{t$_{i-1}$} & - \\ & & \textbf{s$_{i-1: i}$}, \textbf{t$_{0:i-1}$} & - \\ \midrule
G2sD7N53ju8-5 & D & ground truth & If you delete the wrong thing, you can always undo it by pressing Apple Z as well. \\ & & \textbf{s$_i$} & If you do the wrong thing you can always undo it by pressing the apple Z. \\ & & \textbf{s$_{i-1:i}$} & If you do the wrong thing you can always undo it by pressing the apple button and Z. \\ & & \textbf{s$_{i-1:i}$}, \textbf{t$_{i-1}$} & - \\ & & \textbf{s$_{i-1: i}$}, \textbf{t$_{0:i-1}$} & - \\ \midrule
G2uKe6hCNSo-5 & D & ground truth & You can get these mostly at a good paper supply or art supply places will cost you a little bit more, so look for a paper supply. \\ & & \textbf{s$_i$} & A few paper supply or art supply places will charge you a little more so look for paper supplies. \\ & & \textbf{s$_{i-1:i}$} & Got a few good paper supply places - art supply places will charge you a little bit more so look for paper supply. \\ & & \textbf{s$_{i-1:i}$}, \textbf{t$_{i-1}$} & - \\ & & \textbf{s$_{i-1: i}$}, \textbf{t$_{0:i-1}$} & - \\ \midrule
G38DbiHHTW0-5 & D & ground truth & So I'm holding it naturally like I was going to do the basic cradle, right, and I'm just, I'm moving my arms all the way across, so I've got my right arm across my body, I turn my stick out so it's flat and I'm going to pass the ball like that, alright? \\ & & \textbf{s$_i$} & So I'm holding it naturally like I was going to do with the base handle. Right. And I'm going to move it over here. So I have my right arm low and I'm going to raise it so it's in front and then turn it over. I'm going to pass the ball like that, alright? \\ & & \textbf{s$_{i-1:i}$} & - \\ & & \textbf{s$_{i-1:i}$}, \textbf{t$_{i-1}$} & - \\ & & \textbf{s$_{i-1: i}$}, \textbf{t$_{0:i-1}$} & - \\ \midrule
G3CyVk6dizw-5 & D & ground truth & That's basically what we mean when we say we're dubbing the body. \\ & & \textbf{s$_i$} & That's what we mean when we say we are dubbing the body. \\ & & \textbf{s$_{i-1:i}$} & - \\ & & \textbf{s$_{i-1:i}$}, \textbf{t$_{i-1}$} & That's basically what we mean when we say we are dubbing the body. \\ & & \textbf{s$_{i-1: i}$}, \textbf{t$_{0:i-1}$} & - \\ \midrule
G3EE6yhl1vk-5 & D & ground truth & You don't want to hit it to where you restrict it because then, you're definitely going to come up with a cracked cymbal somewhere along the line and if you're paying for them yourselves, you'll understand that a couple hundred of dollars a cymbal is not cheap. \\ & & \textbf{s$_i$} & But you dont want to hit where R-B limit. Why? Because then you are definitely going to come up with a cracked cymbal somewhere on the line. You will understand that's a few hundred dollars of cymbal, it's not cheap. \\ & & \textbf{s$_{i-1:i}$} & But you dont want to hit where the rubber restraint is. Why? Because then you are definitely going to come up with a cracked cymbal somewhere on the line. You will understand that's a few hundred dollars of cymbal, it's not cheap. \\ & & \textbf{s$_{i-1:i}$}, \textbf{t$_{i-1}$} & - \\ & & \textbf{s$_{i-1: i}$}, \textbf{t$_{0:i-1}$} & - \\ \midrule
G3EYpadwqck-5 & D & ground truth & You want to make sure that the liquid is clear and color free. \\ & & \textbf{s$_i$} & I want you to make sure that the liquid is clear and color-free. \\ & & \textbf{s$_{i-1:i}$} & - \\ & & \textbf{s$_{i-1:i}$}, \textbf{t$_{i-1}$} & - \\ & & \textbf{s$_{i-1: i}$}, \textbf{t$_{0:i-1}$} & - \\
        \bottomrule
    \end{tabular}
    \label{tab:baseline6}
\end{table*}

\begin{table*}[]
    \centering
    \small
    \begin{tabular}{cccp{9.2cm}}
        \toprule
        \textbf{video id} & \textbf{interpreter} & \textbf{setting} & \textbf{translation} \\
        \midrule

G3GcPpidwxk-5 & D & ground truth & It always looks like a tuxedo. \\ & & \textbf{s$_i$} & Always looks like a tux. \\ & & \textbf{s$_{i-1:i}$} & - \\ & & \textbf{s$_{i-1:i}$}, \textbf{t$_{i-1}$} & - \\ & & \textbf{s$_{i-1: i}$}, \textbf{t$_{0:i-1}$} & Bow ties will always fit the look of a tuxedo \\ \midrule
G3HKHxevpFI-5 & D & ground truth & Any facial scrub you don't want to use it more than about three times per week. \\ & & \textbf{s$_i$} & You don't want to use that other face scrub more than three times a week. \\ & & \textbf{s$_{i-1:i}$} & You don't want to use a face scrub more than three times a week. \\ & & \textbf{s$_{i-1:i}$}, \textbf{t$_{i-1}$} & - \\ & & \textbf{s$_{i-1: i}$}, \textbf{t$_{0:i-1}$} & - \\ \midrule
G3IJAoK0uSE-5 & D & ground truth & I've used a portion of the back scenery here, a little; just a small clip of the city. \\ & & \textbf{s$_i$} & Have used a portion of the back scenery here, a small metal movie city. \\ & & \textbf{s$_{i-1:i}$} & - \\ & & \textbf{s$_{i-1:i}$}, \textbf{t$_{i-1}$} & - \\ & & \textbf{s$_{i-1: i}$}, \textbf{t$_{0:i-1}$} & - \\ \midrule
G3bMqicS4bQ-5 & D & ground truth & It's got 102 different classes, and this is where you really need to take your specific car, go to the rule book, go to, you know, go online and find out where your car falls in that because that's going to give you your handicap. \\ & & \textbf{s$_i$} & Have 102 different categories and this is where you really have to know your specific car and the rulebook, and find out where your car falls in that because that's going to give you your HC. \\ & & \textbf{s$_{i-1:i}$} & Have 102 different categories and this is where you really have to know your specific car and the rulebook, go online and find out where your car falls in that because that's going to give you your HC. \\ & & \textbf{s$_{i-1:i}$}, \textbf{t$_{i-1}$} & - \\ & & \textbf{s$_{i-1: i}$}, \textbf{t$_{0:i-1}$} & - \\ \midrule
G3g0-BeFN3c-5 & D & ground truth & Any type of modeling. \\ & & \textbf{s$_i$} & All types of models. \\ & & \textbf{s$_{i-1:i}$} & All kinds of modeling. \\ & & \textbf{s$_{i-1:i}$}, \textbf{t$_{i-1}$} & All kinds of modeling. \\ & & \textbf{s$_{i-1: i}$}, \textbf{t$_{0:i-1}$} & - \\ \midrule
G3gm\_C5UueQ-5 & D & ground truth & So, I'm going to turn on my sequencer and I'm just going to press play and you can just go to each one and just hear a different presets. \\ & & \textbf{s$_i$} & I'm going to turn on my sequencer and I'm going to push play and you can say go to each one and listen to different presets \\ & & \textbf{s$_{i-1:i}$} & - \\ & & \textbf{s$_{i-1:i}$}, \textbf{t$_{i-1}$} & - \\ & & \textbf{s$_{i-1: i}$}, \textbf{t$_{0:i-1}$} & - \\ \midrule
G3k86AVFwVs-5 & D & ground truth & If the partial which rests on the tooth, is held up by the plastic portion, or the metal portion, not allowing the partial to completely cede against the tissue. \\ & & \textbf{s$_i$} & If the part which rests on the tooth is held up by the plastic part or the metal part... Not allow the part to be complete ede against the tissue. \\ & & \textbf{s$_{i-1:i}$} & If the part which rests on the tooth is held up by the plastic part or the metal part, it won't allow the part to be completely ceded against the tissue. \\ & & \textbf{s$_{i-1:i}$}, \textbf{t$_{i-1}$} & - \\ & & \textbf{s$_{i-1: i}$}, \textbf{t$_{0:i-1}$} & - \\ \midrule
G3qZW-hZXaQ-5 & D & ground truth & So if someone is coming at you with a knife and they stab straight in it is best to turn out of the way. \\ & & \textbf{s$_i$} & If someone comes close to you with a knife and stabs you directly... \\ & & \textbf{s$_{i-1:i}$} & - \\ & & \textbf{s$_{i-1:i}$}, \textbf{t$_{i-1}$} & - \\ & & \textbf{s$_{i-1: i}$}, \textbf{t$_{0:i-1}$} & - \\ \midrule
\_G0MZFLIHa0-5 & D & ground truth & So first thing's first. \\ & & \textbf{s$_i$} & First things first. \\ & & \textbf{s$_{i-1:i}$} & - \\ & & \textbf{s$_{i-1:i}$}, \textbf{t$_{i-1}$} & - \\ & & \textbf{s$_{i-1: i}$}, \textbf{t$_{0:i-1}$} & - \\ \midrule
\_fZbAxSSbX4-5 & D & ground truth & If you miss one, try to regroup and try to keep throwing. \\ & & \textbf{s$_i$} & If you miss one, try to regroup, and try to keep throwing. \\ & & \textbf{s$_{i-1:i}$} & - \\ & & \textbf{s$_{i-1:i}$}, \textbf{t$_{i-1}$} & - \\ & & \textbf{s$_{i-1: i}$}, \textbf{t$_{0:i-1}$} & - \\

        \bottomrule
    \end{tabular}
    \label{tab:baseline7}
\end{table*}

\begin{table*}[]
    \centering
    \small
    \begin{tabular}{cccp{9.2cm}}
        \toprule
        \textbf{video id} & \textbf{interpreter} & \textbf{setting} & \textbf{translation} \\
        \midrule
fZgWKh3ENoE-8 & C & ground truth & It helps supplements all this, by discarding Destiny Hero Disk Commander to the graveyard with Destiny Draw and then drawing cards. \\ & & \textbf{s$_i$} & Help supplement all of this by discarding DH disk on mine. Crosses on a grave with destiny drov, and drawing cards. \\ & & \textbf{s$_{i-1:i}$} & Help supplement all of this by discarding DH disk on my Commander. Crosses on a grave with destiny drov, and drawing cards. \\ & & \textbf{s$_{i-1:i}$}, \textbf{t$_{i-1}$} & - \\ & & \textbf{s$_{i-1: i}$}, \textbf{t$_{0:i-1}$} & Help supplement all of this by discarding Destiny Hero disk to my graveyard with destiny drov, and drawing cards. \\ \midrule
fZgbCwSG3Hc-8 & C & ground truth & Once again. most creatures in most decks, except for blue, will not come with flying. So, if you are having trouble with flying creatures, you should put a couple Whalebone Gliders in your creature deck. \\ & & \textbf{s$_i$} & Most creatures - most tiles except blue does not come with flying. If you're struggling with flying creatures you should go ahead and add WG on your creature. \\ & & \textbf{s$_{i-1:i}$} & Most creature cards except for blue don't come with flying. If you're struggling with flying creatures you should go ahead and add WG on your creature. \\ & & \textbf{s$_{i-1:i}$}, \textbf{t$_{i-1}$} & Most creature cards except for blue don't come with flying. If you're struggling with flying creatures you should go ahead and add Whalebone Glider on your creature. \\ & & \textbf{s$_{i-1: i}$}, \textbf{t$_{0:i-1}$} & - \\ \midrule
fzXgYPSnaDs-8 & C & ground truth & All you do, you take your maggot, you can use meal worms, as well, which are much bigger, which are probably more well suited for this because this is a rather large hook. \\ & & \textbf{s$_i$} & Okay so what are you all doing now? Maggots or mealworms may be more suited for this since it's large. \\ & & \textbf{s$_{i-1:i}$} & Okay so what are you all doing now? Maggots - or mealworms may be more suited for this since it's a large hook. \\ & & \textbf{s$_{i-1:i}$}, \textbf{t$_{i-1}$} & - \\ & & \textbf{s$_{i-1: i}$}, \textbf{t$_{0:i-1}$} & - \\ \midrule
fzXsxNFczRA-8 & C & ground truth & So that it is a two piece gourd rather than just a simple bowl. \\ & & \textbf{s$_i$} & It's missing something cool, rather than just having the bowl. \\ & & \textbf{s$_{i-1:i}$} & - parts, very cool. Better than just having the bowl. \\ & & \textbf{s$_{i-1:i}$}, \textbf{t$_{i-1}$} & - \\ & & \textbf{s$_{i-1: i}$}, \textbf{t$_{0:i-1}$} & - \\ \midrule
fzcsY2gm7t0-8 & C & ground truth & Composition is what is going to control the flow of the viewer's experience in the space. \\ & & \textbf{s$_i$} & What controls the flow of the viewing experience. \\ & & \textbf{s$_{i-1:i}$} & - \\ & & \textbf{s$_{i-1:i}$}, \textbf{t$_{i-1}$} & - \\ & & \textbf{s$_{i-1: i}$}, \textbf{t$_{0:i-1}$} & - \\ \midrule
fzncPNr2Sc0-8 & C & ground truth & You'll click on that and then they'll want you to sign in and the first time that you try to do it, there's a process of signing in, and creating a password. \\ & & \textbf{s$_i$} & Touch the button and a window will pop up for you to sign in. If it's the first time you'll have to go through the process of setting up a username and password. \\ & & \textbf{s$_{i-1:i}$} & - \\ & & \textbf{s$_{i-1:i}$}, \textbf{t$_{i-1}$} & - \\ & & \textbf{s$_{i-1: i}$}, \textbf{t$_{0:i-1}$} & - \\ \midrule
g05yGRoZE10-8 & C & ground truth & This one's very nicely used. \\ & & \textbf{s$_i$} & Kind of used often. \\ & & \textbf{s$_{i-1:i}$} & Already well-used. \\ & & \textbf{s$_{i-1:i}$}, \textbf{t$_{i-1}$} & - \\ & & \textbf{s$_{i-1: i}$}, \textbf{t$_{0:i-1}$} & - \\ \midrule
g0S7FAqIweA-8 & C & ground truth & To place the waist strap in place, we snap the buckle, and locate the ends of the strap, to tighten the SCBA unit, on to the waist. \\ & & \textbf{s$_i$} & Buckle the seatbelt and tighten it. \\ & & \textbf{s$_{i-1:i}$} & - \\ & & \textbf{s$_{i-1:i}$}, \textbf{t$_{i-1}$} & Buckle the waist strap and tighten it. \\ & & \textbf{s$_{i-1: i}$}, \textbf{t$_{0:i-1}$} & - \\
        \bottomrule
    \end{tabular}
    \label{tab:baseline8}
\end{table*}

\begin{table*}[]
    \centering
    \small
    \begin{tabular}{cccp{9.2cm}}
        \toprule
        \textbf{video id} & \textbf{interpreter} & \textbf{setting} & \textbf{translation} \\
        \midrule

g0TkUiO7t4I-8 & C & ground truth & Cause wrinkle is the problem with the dry skin. \\ & & \textbf{s$_i$} & The rash is a problem with dry skin. \\ & & \textbf{s$_{i-1:i}$} & - \\ & & \textbf{s$_{i-1:i}$}, \textbf{t$_{i-1}$} & Wrinkles are a problem with dry skin. \\ & & \textbf{s$_{i-1: i}$}, \textbf{t$_{0:i-1}$} & - \\ \midrule
g0fgci8L\_rc-8 & C & ground truth & No big deal. \\ & & \textbf{s$_i$} & Water. \\ & & \textbf{s$_{i-1:i}$} & - \\ & & \textbf{s$_{i-1:i}$}, \textbf{t$_{i-1}$} & - \\ & & \textbf{s$_{i-1: i}$}, \textbf{t$_{0:i-1}$} & - \\ \midrule
g0iNy-yPisM-8 & C & ground truth & So, when you've completed making all of your edits, and mind you, you can use HTML, if you like. \\ & & \textbf{s$_i$} & When you're finished editing you can use HTML if you want \\ & & \textbf{s$_{i-1:i}$} & - \\ & & \textbf{s$_{i-1:i}$}, \textbf{t$_{i-1}$} & - \\ & & \textbf{s$_{i-1: i}$}, \textbf{t$_{0:i-1}$} & - \\ \midrule
g0pRnlPR-K0-8 & C & ground truth & Right now we're discussing setting the water level on your machine. \\ & & \textbf{s$_i$} & Right now I'm discussing setting up the water level of your machine. \\ & & \textbf{s$_{i-1:i}$} & - \\ & & \textbf{s$_{i-1:i}$}, \textbf{t$_{i-1}$} & - \\ & & \textbf{s$_{i-1: i}$}, \textbf{t$_{0:i-1}$} & - \\ \midrule
g0t4Wz5qsT8-8 & C & ground truth & So, for the sake of comedy let's see what happens. \\ & & \textbf{s$_i$} & For the goal of it, I'll go ahead and show you, see what happens \\ & & \textbf{s$_{i-1:i}$} & - \\ & & \textbf{s$_{i-1:i}$}, \textbf{t$_{i-1}$} & - \\ & & \textbf{s$_{i-1: i}$}, \textbf{t$_{0:i-1}$} & - \\ \midrule
g1HXoDkax5A-3 & E & ground truth & I could try to drive up the nose; it's very effective. \\ & & \textbf{s$_i$} & Up the nose. \\ & & \textbf{s$_{i-1:i}$} & Undercut and strike up at the nose. \\ & & \textbf{s$_{i-1:i}$}, \textbf{t$_{i-1}$} & - \\ & & \textbf{s$_{i-1: i}$}, \textbf{t$_{0:i-1}$} & - \\ \midrule
g1HvmBOR7Y4-3 & E & ground truth & Put it over their head, give them a treat. \\ & & \textbf{s$_i$} & Put the collar on then give them treats. \\ & & \textbf{s$_{i-1:i}$} & - \\ & & \textbf{s$_{i-1:i}$}, \textbf{t$_{i-1}$} & - \\ & & \textbf{s$_{i-1: i}$}, \textbf{t$_{0:i-1}$} & - \\ \midrule
g1uA0f9I0Sg-3 & E & ground truth & If you are looking to buy hosiery for open toe shoes, be it if they are peep toe shoes or if you are looking to wear hosiery with a sandal in the wintertime your best options are to go with hosiery that doesn't have any hem lines or any type of reinforcements. \\ & & \textbf{s$_i$} & If you want peep toe shoes or sandals in winter, you should still pick hoir with no lines or reinforcement \\ & & \textbf{s$_{i-1:i}$} & Whether you want to wear open toed shoes like sandals or winter shoes, you should still pick hosery with no lines or reinforcement \\ & & \textbf{s$_{i-1:i}$}, \textbf{t$_{i-1}$} & Whether you want to wear open toed shoes like sandals or winter shoes, you should still pick hosiery with no lines or reinforcement \\ & & \textbf{s$_{i-1: i}$}, \textbf{t$_{0:i-1}$} & - \\ \midrule
g1vUH8Iy4vw-3 & E & ground truth & We'll start with your feet comfortable, a little wider than your hips maybe. \\ & & \textbf{s$_i$} & Start with your feet comfortable, a little wider than your hips. \\ & & \textbf{s$_{i-1:i}$} & - \\ & & \textbf{s$_{i-1:i}$}, \textbf{t$_{i-1}$} & - \\ & & \textbf{s$_{i-1: i}$}, \textbf{t$_{0:i-1}$} & - \\ \midrule
g1xdqxCZxTg-3 & E & ground truth & I thing that would be awesome. \\ & & \textbf{s$_i$} & Calm down. \\ & & \textbf{s$_{i-1:i}$} & Whoa. \\ & & \textbf{s$_{i-1:i}$}, \textbf{t$_{i-1}$} & - \\ & & \textbf{s$_{i-1: i}$}, \textbf{t$_{0:i-1}$} & - \\ \midrule
g1z6HOJ0yRw-3 & E & ground truth & They're just supporting me. \\ & & \textbf{s$_i$} & Support alone. \\ & & \textbf{s$_{i-1:i}$} & Support only. \\ & & \textbf{s$_{i-1:i}$}, \textbf{t$_{i-1}$} & - \\ & & \textbf{s$_{i-1: i}$}, \textbf{t$_{0:i-1}$} & - \\

        \bottomrule
    \end{tabular}
    \label{tab:baseline9}
\end{table*}

\begin{table*}[]
    \centering
    \small
    \begin{tabular}{cccp{9.2cm}}
        \toprule
        \textbf{video id} & \textbf{interpreter} & \textbf{setting} & \textbf{translation} \\
        \midrule

g2NA\_eBUcH8-3 & E & ground truth & You're not playing with any other player. \\ & & \textbf{s$_i$} & Not against any of them. \\ & & \textbf{s$_{i-1:i}$} & - \\ & & \textbf{s$_{i-1:i}$}, \textbf{t$_{i-1}$} & - \\ & & \textbf{s$_{i-1: i}$}, \textbf{t$_{0:i-1}$} & - \\ \midrule
g2QdwYqm8pg-3 & E & ground truth & This time it is going to be a white face. \\ & & \textbf{s$_i$} & Now white face. \\ & & \textbf{s$_{i-1:i}$} & Now my face is white. \\ & & \textbf{s$_{i-1:i}$}, \textbf{t$_{i-1}$} & - \\ & & \textbf{s$_{i-1: i}$}, \textbf{t$_{0:i-1}$} & - \\ \midrule
g2SdWBPoXZ0-3 & E & ground truth & So, for example, if I'm going to build up into a backcross pattern, I don't want to just go here and immediately start throwing backcrosses. \\ & & \textbf{s$_i$} & For example, if I'm building to a back cross pattern, I don't want to just go along and then back cross. \\ & & \textbf{s$_{i-1:i}$} & - \\ & & \textbf{s$_{i-1:i}$}, \textbf{t$_{i-1}$} & - \\ & & \textbf{s$_{i-1: i}$}, \textbf{t$_{0:i-1}$} & - \\ \midrule
g2eTD-1Jcro-3 & E & ground truth & To do the butterfly breath flow, it helps you think about the alignment and also makes you think about your breath. \\ & & \textbf{s$_i$} & Straighten your spine and breathe. \\ & & \textbf{s$_{i-1:i}$} & - \\ & & \textbf{s$_{i-1:i}$}, \textbf{t$_{i-1}$} & - \\ & & \textbf{s$_{i-1: i}$}, \textbf{t$_{0:i-1}$} & - \\ \midrule
g2iFC1st7zQ-3 & E & ground truth & And there you go. \\ & & \textbf{s$_i$} & There you go. \\ & & \textbf{s$_{i-1:i}$} & - \\ & & \textbf{s$_{i-1:i}$}, \textbf{t$_{i-1}$} & - \\ & & \textbf{s$_{i-1: i}$}, \textbf{t$_{0:i-1}$} & - \\ \midrule
g2nvBjp0loQ-3 & E & ground truth & Reach up to the fingers, to the side, I've got my sides here, this way, then the lower back. \\ & & \textbf{s$_i$} & Up, fingers, this side, other side, this way, then lower back. \\ & & \textbf{s$_{i-1:i}$} & First stretch your arms and fingers up and to the side, other side, this way. Then lower back. \\ & & \textbf{s$_{i-1:i}$}, \textbf{t$_{i-1}$} & - \\ & & \textbf{s$_{i-1: i}$}, \textbf{t$_{0:i-1}$} & - \\ \midrule
g2o-GFdGOJE-3 & E & ground truth & Probably not going to catch a flush with a three to it. \\ & & \textbf{s$_i$} & Not getting a flush with three. \\ & & \textbf{s$_{i-1:i}$} & - \\ & & \textbf{s$_{i-1:i}$}, \textbf{t$_{i-1}$} & - \\ & & \textbf{s$_{i-1: i}$}, \textbf{t$_{0:i-1}$} & - \\ \midrule
g2v-M6EXcUE-3 & E & ground truth & Basil is best harvested when there's a lot of leafy stuff right at the tip, but not a flowering stalk yet. \\ & & \textbf{s$_i$} & Best to harvest when there's a lot of leaves on the top but not yet any flowers. \\ & & \textbf{s$_{i-1:i}$} & - \\ & & \textbf{s$_{i-1:i}$}, \textbf{t$_{i-1}$} & - \\ & & \textbf{s$_{i-1: i}$}, \textbf{t$_{0:i-1}$} & Basil is best to harvest when there's a lot of leaves on the top but not yet any flowers. \\ \midrule
g38AmwPAYvg-3 & E & ground truth & And, take them out and take some pictures of them in the sunlight and see how the sun reflects on their skin and how the camera reacts with that, then grab a white piece of paper and hold it up and reflect the light back onto their skin. \\ & & \textbf{s$_i$} & Go outside and take some pictures with the sun. See how the sun reflects on skin and how the camera reacts to that. Then get a white paper, hold it up, and reflect light back on the skin. \\ & & \textbf{s$_{i-1:i}$} & - \\ & & \textbf{s$_{i-1:i}$}, \textbf{t$_{i-1}$} & - \\ & & \textbf{s$_{i-1: i}$}, \textbf{t$_{0:i-1}$} & - \\ \midrule
g3Cc\_1-V31U-3 & E & ground truth & I kind of like that. \\ & & \textbf{s$_i$} & You love me? OK. \\ & & \textbf{s$_{i-1:i}$} & - \\ & & \textbf{s$_{i-1:i}$}, \textbf{t$_{i-1}$} & - \\ & & \textbf{s$_{i-1: i}$}, \textbf{t$_{0:i-1}$} & - \\

        \bottomrule
    \end{tabular}
    \label{tab:baseline10}
\end{table*}

\begin{table*}[]
    \centering
    \small
    \begin{tabular}{cccp{9.2cm}}
        \toprule
        \textbf{video id} & \textbf{interpreter} & \textbf{setting} & \textbf{translation} \\
        \midrule
g3DkYITeIy0-3 & E & ground truth & The custom trays are wonderful. \\ & & \textbf{s$_i$} & Wonderful. \\ & & \textbf{s$_{i-1:i}$} & - \\ & & \textbf{s$_{i-1:i}$}, \textbf{t$_{i-1}$} & - \\ & & \textbf{s$_{i-1: i}$}, \textbf{t$_{0:i-1}$} & - \\ \midrule
g3PBeTb1TCw-3 & E & ground truth & So one more time. \\ & & \textbf{s$_i$} & One more time. \\ & & \textbf{s$_{i-1:i}$} & - \\ & & \textbf{s$_{i-1:i}$}, \textbf{t$_{i-1}$} & - \\ & & \textbf{s$_{i-1: i}$}, \textbf{t$_{0:i-1}$} & - \\ \midrule
g3V0BsmDUgY-3 & E & ground truth & My name is Sylvia Russell and this is how you choose a hair style for your face shape. \\ & & \textbf{s$_i$} & My name is Sylvia Russel and that's how to pick a hairstyle for your face shape. \\ & & \textbf{s$_{i-1:i}$} & - \\ & & \textbf{s$_{i-1:i}$}, \textbf{t$_{i-1}$} & - \\ & & \textbf{s$_{i-1: i}$}, \textbf{t$_{0:i-1}$} & My name is Sylvia Russell and that's how to pick a hairstyle for your face shape. \\ \midrule
g3X3XE6M2\_A-3 & E & ground truth & This is where he's strong. \\ & & \textbf{s$_i$} & That's where strong. \\ & & \textbf{s$_{i-1:i}$} & - \\ & & \textbf{s$_{i-1:i}$}, \textbf{t$_{i-1}$} & That's where strength. \\ & & \textbf{s$_{i-1: i}$}, \textbf{t$_{0:i-1}$} & - \\ \midrule
g3ZgF8gdfLo-3 & E & ground truth & You then add the top of the condenser, which fastens on with three clips, or clamps. \\ & & \textbf{s$_i$} & Then you add the lid and clip the three latches on. \\ & & \textbf{s$_{i-1:i}$} & - \\ & & \textbf{s$_{i-1:i}$}, \textbf{t$_{i-1}$} & - \\ & & \textbf{s$_{i-1: i}$}, \textbf{t$_{0:i-1}$} & - \\ \midrule
g3jQ5ecjGz8-3 & E & ground truth & So I'll just go ahead and use the pistol that we picked up from the gangster downstairs, and shoot the chemist. \\ & & \textbf{s$_i$} & I picked up this gun from the gangster downstairs. They are shooting chemists! \\ & & \textbf{s$_{i-1:i}$} & I pick up this pistol from the gangster downstairs, then I shoot the chemist! \\ & & \textbf{s$_{i-1:i}$}, \textbf{t$_{i-1}$} & The pistol that I picked up from the gangster downstairs. Then I shoot the chemist! \\ & & \textbf{s$_{i-1: i}$}, \textbf{t$_{0:i-1}$} & - \\ \midrule
g3kFAmcBpFc-3 & E & ground truth & This is a shampoo by Verback. \\ & & \textbf{s$_i$} & This shampoo from Verback. \\ & & \textbf{s$_{i-1:i}$} & - \\ & & \textbf{s$_{i-1:i}$}, \textbf{t$_{i-1}$} & - \\ & & \textbf{s$_{i-1: i}$}, \textbf{t$_{0:i-1}$} & - \\ \midrule
g3pXM5X3\_Xw-3 & E & ground truth & Needle tool. \\ & & \textbf{s$_i$} & Needle tool. \\ & & \textbf{s$_{i-1:i}$} & - \\ & & \textbf{s$_{i-1:i}$}, \textbf{t$_{i-1}$} & - \\ & & \textbf{s$_{i-1: i}$}, \textbf{t$_{0:i-1}$} & - \\ \midrule
g3sLd8JupoQ-3 & E & ground truth & I'll measure down two inches and put a mark, and then on two inches on the other side and put a mark. \\ & & \textbf{s$_i$} & Measure 2 inches then mark it. Then 2 inches on the other side and mark. \\ & & \textbf{s$_{i-1:i}$} & - \\ & & \textbf{s$_{i-1:i}$}, \textbf{t$_{i-1}$} & - \\ & & \textbf{s$_{i-1: i}$}, \textbf{t$_{0:i-1}$} & - \\ \midrule
g3ushtMfLiY-3 & E & ground truth & In order to have your veil in the middle of the choreography, before you get on stage you are going to get your veil and you are going to place it on your hips like this. \\ & & \textbf{s$_i$} & If you want a veil in the middle of the show/dance, before you arrive on stage, get your veil and put it on your hips like that. \\ & & \textbf{s$_{i-1:i}$} & - \\ & & \textbf{s$_{i-1:i}$}, \textbf{t$_{i-1}$} & - \\ & & \textbf{s$_{i-1: i}$}, \textbf{t$_{0:i-1}$} & - \\

        \bottomrule
    \end{tabular}
    \label{tab:baseline11}
\end{table*}

\end{document}